\documentclass[sigconf]{acmart}

\usepackage{wrapfig}
\usepackage{amsmath}
\PassOptionsToPackage{numbers}{natbib}
\usepackage{natbib}
\usepackage{multirow}
\usepackage{booktabs}
\usepackage{hyperref}
\usepackage{diagbox}
\usepackage{multicol}
\hypersetup{
    colorlinks=true,
    linkcolor=blue,
    filecolor=magenta,
    urlcolor=magenta,
}
\usepackage[linesnumbered,algoruled,boxed,lined, noend]{algorithm2e}
\usepackage{algpseudocode}
\usepackage{makecell}
\usepackage[all]{hypcap}
\usepackage{booktabs}
\usepackage{url}
\usepackage{xcolor}
\usepackage{tikz}
\usepackage{diagbox}
\usepackage{listings}
\usepackage{comment} 
\usepackage{graphicx}
\usepackage{longtable}
\usepackage{array}
\usepackage{colortbl}
\usepackage{caption}

\theoremstyle{definition}

\usepackage{tcolorbox}

\tcbset{
    colback=gray!10, 
    colframe=gray, 
    coltext=black, 
    boxsep=5pt, 
    arc=0mm, 
    top=0mm, bottom=0mm, 
    left=1mm, right=1mm, 
    boxrule=0.5mm, 
    toptitle=1mm, bottomtitle=1mm, 
    titlerule=0mm, 
}
\definecolor{deepgreen}{rgb}{0.0, 0.4, 0.0}

\definecolor{codegreen}{rgb}{0,0.6,0}
\definecolor{codegray}{rgb}{0.5,0.5,0.5}
\definecolor{codepurple}{rgb}{0.58,0,0.82}
\definecolor{backcolour}{rgb}{0.95,0.95,0.92}

\lstdefinestyle{mystyle}{
  backgroundcolor=\color{backcolour}, commentstyle=\color{codegreen},
  keywordstyle=\color{magenta},
  numberstyle=\tiny\color{codegray},
  stringstyle=\color{codepurple},
  basicstyle=\ttfamily\footnotesize,
  breakatwhitespace=false,         
  breaklines=true,                 
  captionpos=b,                    
  keepspaces=true,                 
  numbers=left,                    
  numbersep=5pt,                  
  showspaces=false,                
  showstringspaces=false,
  showtabs=false,                  
  tabsize=2
}
\usepackage{xurl}

\newcommand{\find}[1]{
\begin{tcolorbox}[leftrule=0.5mm,toprule=0mm,bottomrule=0mm,left=0.7pt,right=0.7pt,top=0.2pt,bottom=0.2pt]
\em #1
\end{tcolorbox}
}

\usepackage[framemethod=TikZ]{mdframed}
\definecolor{mycolor}{RGB}{194, 214, 236}

\newcounter{result}

\makeatletter
\g@addto@macro{\@algocf@init}{\SetKwInOut{Parameter}{Parameters}}
\makeatother

\lstdefinestyle{yaml}{
     basicstyle=\color{blue}\tiny,
     rulecolor=\color{black},
     string=[s]{'}{'},
     stringstyle=\color{blue},
     comment=[l]{:},
     commentstyle=\color{black},
     morecomment=[l]{-}
 }

\AtBeginDocument{%
  }

\setcopyright{none}
\copyrightyear{2026}

\begin{document}

\date{}

\title{Implicit Patterns in LLM-Based Binary Analysis}
\thanks{Artifacts: \url{https://github.com/bjtu-SecurityLab/pattern-tl}}
\author{Qiang Li}
\affiliation{
  \institution{Beijing Jiaotong University}
  \country{China}
}
\email{liqiang@bjtu.edu.cn}
\orcid{0000-0001-9833-2836}  

\author{XiangRui Zhang}
\affiliation{
  \institution{Beijing Jiaotong University}
  \country{China}
}
\orcid{0009-0002-9124-6665}  

\author{Haining Wang}
\affiliation{
  \institution{Virginia Tech}
  \country{USA}
}
\email{hnw@vt.edu}
\orcid{0000-0002-4174-3009}  


\begin{abstract}
Binary vulnerability analysis is increasingly performed by LLM-based agents in an iterative, multi-pass manner, with the model as the core decision-maker.
However, how such systems organize exploration over hundreds of reasoning steps remains poorly understood, due to limited context windows and implicit token-level behaviors.
We present the first large-scale, trace-level study showing that multi-pass LLM reasoning gives rise to structured, token-level implicit patterns.
Analyzing 521 binaries with 99,563 reasoning steps, we identify four dominant patterns—early pruning, path-dependent lock-in, targeted backtracking, and knowledge-guided prioritization—that emerge implicitly from reasoning traces.
These token-level implicit patterns serve as an abstraction of LLM reasoning: instead of explicit control-flow or predefined heuristics, exploration is organized through implicit decisions regulating path selection, commitment, and revision.
Our analysis shows these patterns form a stable, structured system with distinct temporal roles and measurable characteristics.
Our results provide the first systematic characterization of LLM-driven binary analysis and a foundation for more reliable analysis systems.
\end{abstract}


\maketitle

\section{Introduction}


Binary vulnerability analysis~\cite{poeplau2020symbolic,vadayath2022arbiter,gao2024faster,zhao2024leveraging, gibbs2024operation} fundamentally concerns identifying and reasoning about execution paths that connect attacker-influenced sources to security-critical sinks under severe uncertainty.
In practice, such analysis follows two distinct paradigms: \textit{one-pass analysis} and \textit{iterative analysis}.
In one-pass analysis, static analysis constructs a global program representation once, and reasoning operates over this fixed view~\cite{hussain2025vulbinllm,liu2025llm,chen2025clearagent}.
In contrast, iterative analysis interleaves reasoning and tool interaction, where each decision is conditioned on prior observations, reflecting how human analysts explore binaries in practice.
This distinction is fundamental: iterative analysis transforms vulnerability discovery from a static pipeline into a sequential, decision-driven process.

Recent advances in large language models (LLMs) have enabled their application to binary analysis tasks, including disassembly navigation, code reasoning, and interactive, tool-driven exploration~\cite{feng2023prompting,li2023hitchhiker,pearce2023examining,deng2023large,wang2024sanitizing,liu2025llm}.
Crucially, LLMs enable the automation of iterative analysis by acting as the core decision-maker in the analysis loop.
At each step, the model invokes analysis tools and decides which paths to pursue, defer, or abandon—decisions that directly determine whether vulnerability-relevant behaviors are observed.
As a result, the iterative mode, previously requiring human expertise, becomes automatable. 
This shift from one-pass to multi-pass reasoning fundamentally changes how static analysis participates in vulnerability discovery, raising a key question: how do LLM agents organize and sustain exploration under such settings?

However, despite growing interest, little is known about \emph{how} LLM-based systems actually organize binary path exploration.
This gap arises because understanding LLM-driven iterative analysis is fundamentally more challenging than analyzing traditional one-pass systems.
First, agents operate over long analysis sessions spanning hundreds of reasoning steps, where limited context windows and attention drift lead to fragmented observations rather than a coherent global view.
Second, LLM reasoning operates at the token level: behaviors such as prioritization, pruning, and backtracking are not explicitly encoded as program logic or control structures, but emerge implicitly from sequential tokens.
This contrasts with traditional analysis systems, where such behaviors are explicitly defined and enforced.
As a result, it remains unclear whether LLM agents exhibit systematic, structured reasoning behaviors, or merely produce ad hoc decisions tied to specific inputs.

To address this gap, we show that multi-pass LLM reasoning gives rise to structured, token-level implicit patterns that govern long-horizon binary analysis.
We present the first large-scale, trace-level empirical study of how LLM-based systems explore binary execution paths in vulnerability analysis.
We analyze reasoning traces from 521 real-world binaries across ARM and MIPS architectures, comprising 99,563 reasoning steps.
From these traces, we identify four dominant patterns—early pruning, path-dependent lock-in, targeted backtracking, and knowledge-guided prioritization—that emerge \emph{implicitly} from token-level sequential reasoning rather than being externally prescribed by prompts or program logic.
These token-level implicit patterns serve as an abstraction of LLM reasoning: instead of exhaustive enumeration, explicit control-flow traversal, or predefined heuristics, exploration is organized through implicit, token-level decisions that regulate path selection, commitment, and revision over time.

Our analysis reveals three key properties of these patterns.
First, they are stable and repeatedly invoked across sessions rather than sporadic behaviors:
path lock-in and knowledge-guided prioritization appear in 97.6\% of sessions, early pruning in 83.5\%, and backtracking in 93.8\%.
Moreover, they exhibit distinct temporal roles, with lock-in dominating early phases, pruning emerging mid-process, and backtracking concentrated in later stages.
Second, the patterns form structured relationships rather than operating independently.
In particular, path lock-in and early pruning form a bidirectional loop accounting for 79.4\% of pattern transitions, while lock-in and prioritization exhibit complementary dynamics.
Third, each pattern corresponds to distinct, quantifiable behavioral characteristics, as reflected in differences in path length, branching behavior, and backtracking dynamics.
Together, these results show that LLM-driven binary analysis is governed by structured reasoning mechanisms rather than ad hoc exploration.

These findings have implications for both security analysis and LLM-based systems.
For binary analysis, they demonstrate that vulnerability-oriented exploration can emerge without explicit control-flow reconstruction, through iterative, semantic-guided reasoning.
For system design, they provide concrete behavioral primitives that can inform more controllable and effective LLM-based analysis frameworks.

The main contributions of this paper are as follows:
\begin{itemize}
    \item We present the first trace-level empirical study of \emph{long-horizon, iterative} LLM-driven binary analysis, focusing on how agents organize exploration through multi-pass reasoning rather than one-pass execution.

    \item We introduce an \textit{iterative perspective} that reframes binary analysis as a sequential decision process, contrasting with the one-pass paradigm in both traditional static analysis and prior LLM-based approaches.

    \item We identify four recurring \emph{token-level} implicit patterns as an abstraction of LLM reasoning beyond explicit program structures or predefined heuristics.

    \item We show that these patterns have distinct roles, temporal dynamics, and measurable behavioral characteristics, revealing systematic reasoning mechanisms underlying LLM-based binary exploration.
\end{itemize}


\section{Background \& Motivation}
\label{sec:back}

\subsection{One Time or Multiple Times?}

\begin{figure}[!t]
  \centering
  \includegraphics[width=2.6in]{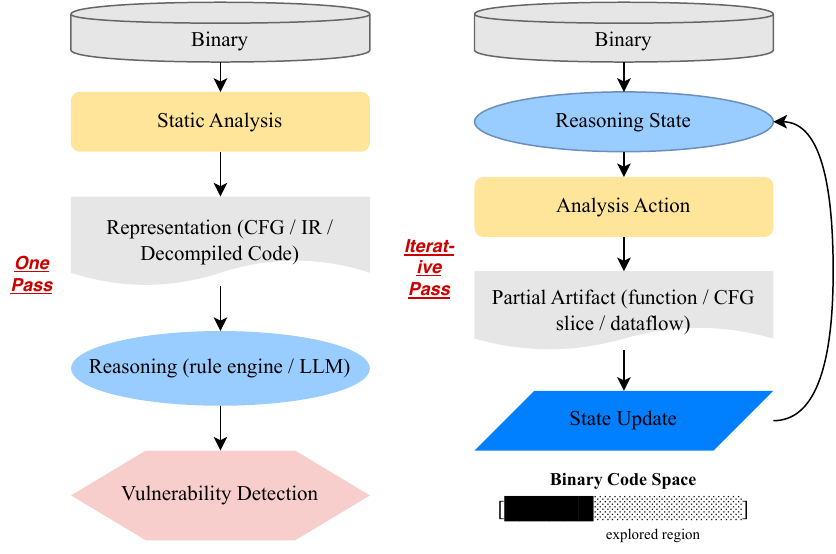}
  \caption{(Left) One-pass paradigm: static analysis constructs a global program representation and vulnerability reasoning operates over this fixed view.
  (Right) Iterative paradigm: reasoning interleaves with repeated static analysis operations, proceeding incrementally through multiple tool invocations.}
  \label{fig:para}
\end{figure}

Binary vulnerability detection can follow two fundamentally different execution paradigms, illustrated in Figure~\ref{fig:para}: \textit{single-pass analysis} and \textit{iterative analysis}.  
The key difference between these paradigms lies in the \emph{role of static analysis} within the vulnerability detection process.

\textbf{One-pass mode.} 
Traditional static analysis tools—disassemblers, decompilers, CFG builders—run once to produce a program representation. 
Subsequent reasoning, whether rule-based, symbolic, or LLM-assisted operates over this artifact. 
The pipeline is: \{ Binary $\to$ Static analysis $\to$ Representation $\to$ Slicing $\to$ Reasoning \}. 
Existing LLM-based binary analysis work follows this structure~\cite{hussain2025vulbinllm,liu2025llm,chen2025clearagent}: the LLM receives a slice of representation (e.g., disassembly, decompiled code) and reasons over it; static analysis remains a one-time preprocessing step producing a fixed representation.

\textbf{Iterative mode.} 
In practice, expert analysts rarely follow a strict one-pass workflow. Analysts often refine their understanding of a binary through additional analysis steps 
during investigation.
They repeatedly invoke static analysis: inspect a function, decide what to examine next based on the result, inspect another region, and so on. 
Reasoning and static analysis are interleaved throughout:
\begin{verbatim}
Binary
 → Reasoning
 → Static analysis
 → Observation
 → Reasoning (loop)
\end{verbatim}
Each invocation (e.g., disassembling a function, resolving cross-references) is a static analysis operation; reasoning emerges incrementally, guided by the preceding observation at each step. 
This iterative mode is how experts actually work; it has not been systematically studied in automated analysis.

\subsection{LLM-Driven Iterative Loop}

\begin{figure*}[!t]
    \centering
    \includegraphics[width=0.95\textwidth]{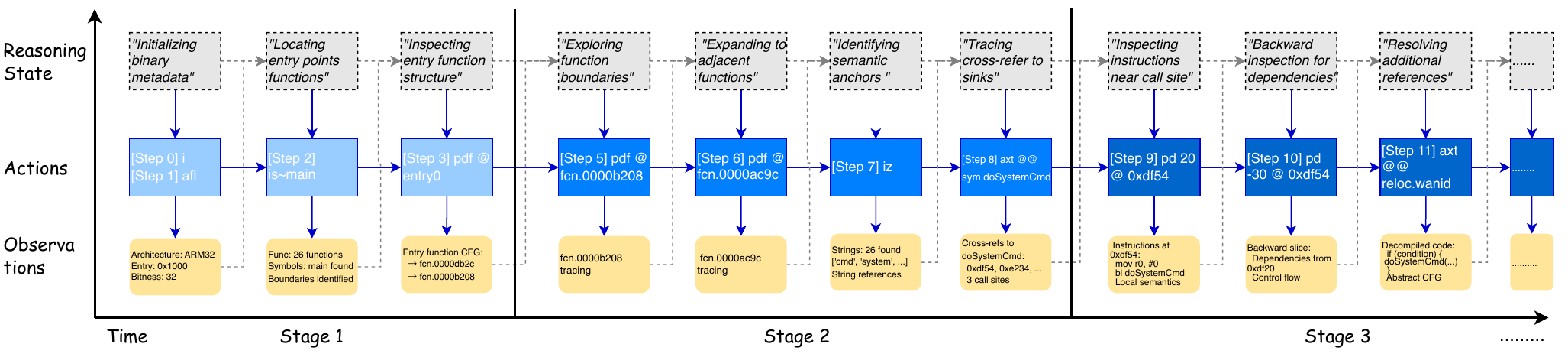}
    \caption{An example of step-by-step LLM-driven binary analysis,
    illustrating how reasoning states evolve through repeated interactions
    with external analysis tools.}
    \label{fig:back-overview}
\end{figure*}

LLM agents that interact with binary analysis tools through tool invocation form a \textit{reasoning–action–observation} loop. 
We focus on agents where the LLM is the sole decision-maker—no external search algorithm or program orchestrates exploration; control emerges from the token-level reasoning that drives each tool invocation.
At each step, the agent selects an analysis command (e.g., disassemble a function, resolve cross-references); the tool returns a localized program fragment; the agent incorporates this into its reasoning and selects the next command. 
This loop is a concrete instantiation of the iterative mode: each action is a static analysis operation, and agents perform hundreds of them in sequence. 
LLM agents thus automate the iterative workflow that experts perform manually, producing observable traces that record how analysis unfolds over time.

Figure~\ref{fig:back-overview} illustrates this process. 
With only a fragment exposed per step, the agent has no global view; it must repeatedly decide what to analyze next, which path to pursue, and which to abandon. The trace—an ordered sequence of reasoning steps and analysis actions—captures these decisions. Such traces exhibit strong temporal dependencies (early decisions shape later exploration) and differ fundamentally from chain-of-thought reasoning~\cite{yao2023tree, wei2022chain, yao2022react, wang2023voyager}: they are grounded in concrete tool invocations and observable program states, not purely internal language. 
In this work, we treat these traces as the primary object of study and analyze token-level implicit patterns that emerge when agents conduct static analysis under partial observability and bounded resources.

\section{Problem Formulation}
\label{sec:problem}

LLM-based binary analysis agents interact with executables through long sequences of tool invocations, observations, and intermediate reasoning steps.
These interactions form long-horizon traces in which exploration control emerges implicitly from token-level reasoning rather than explicit program logic.
This work studies the following problem:

\noindent $\bullet$\textit{What recurring token-level implicit patterns emerge in long-horizon traces produced by LLM agents?}


\subsection{Trace Representation}

We consider an analysis session as a complete interaction between an LLM agent and a target binary, starting from initial reconnaissance and ending when the agent terminates or reports findings. Each session produces an ordered trace
$T = \langle s_1, s_2, \ldots, s_T \rangle$,
where $s_i$ corresponds to a step in the LLM-driven binary analysis process.
Each step $s_i = (a_i, o_i, r_i)$ consists of the issued tool command $a_i$, the tool output $o_i$, and the reasoning state $r_i$ maintained by the agent up to step $i$.
This representation captures how decisions evolve over time and depend on earlier observations and reasoning.
In practice, traces span hundreds of steps and frequently revisit the same code regions, creating long-range dependencies between early decisions and later actions.

\subsection{Token-Level Implicit Behaviors}

In single-pass binary analysis (Section~\ref{sec:back}), whether rule-based, symbolic, or LLM-assisted, behaviors such as prioritization, pruning, focus, and backtracking are explicit: priority queues, branch cuts, deliberate path selection, and stack-based backtracking are first-class operations. 
In LLM agents where the model is the core decision-maker, these behaviors become \emph{implicit}: the agent operates at the token level with no explicit data structures for active paths or deferred candidates; prioritization, pruning, and backtracking manifest in the token stream and tool-invocation sequence. 
Exploration control is thus not directly observable as program structure, but must be inferred from trace dynamics. Extracting and characterizing these behaviors from traces is essential for understanding LLM-based analysis systems, yet they have not been systematically studied in this form.

\subsection{Extracting Token-Level Patterns from Traces}

Our goal is not to assess whether an individual analysis session succeeds, nor to measure vulnerability coverage.
Instead, we aim to characterize the implicit structural mechanisms by which LLM agents organize exploration over time.
Specifically, we study recurring behaviors related to path selection, persistence, revisiting, and backtracking—mechanisms that are not explicitly encoded, but emerge from token-level reasoning dynamics and can be identified from observable trace structure.
By identifying these patterns consistently across binaries and sessions, we reveal organizational principles of LLM-driven binary exploration that are not captured by outcome-based metrics.

\begin{figure}[!t]
    \centering
    \includegraphics[width=0.9\linewidth]{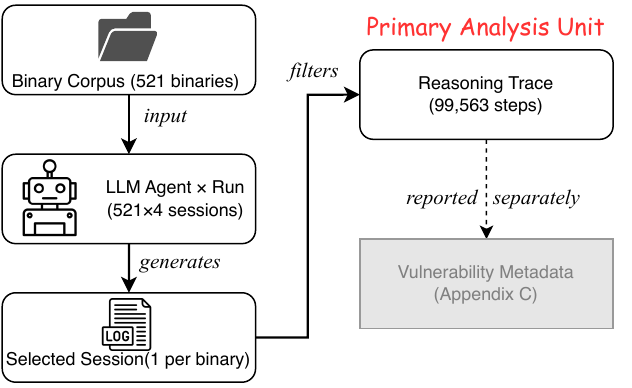}
    \caption{Dataset overview: the relationship between the binary corpus, analysis sessions, reasoning traces, and output metadata.}
    \label{fig:trace-hierarchy}
\end{figure}

\section{Dataset}
\label{sec:dataset}

Our dataset consists of long-horizon reasoning traces collected from LLM-driven binary analysis sessions.
Unlike datasets that focus on binary coverage or vulnerability labels, ours captures the temporal structure of reasoning traces for agent behavior analysis.
We treat the \emph{analysis session} as the primary unit of analysis, rather than the binary itself.
Each session corresponds to one complete interaction between an LLM agent and a target binary, producing a temporally ordered trace of reasoning--action--observation steps.
Figure~\ref{fig:trace-hierarchy} illustrates the hierarchical relationship between the binary corpus, analysis sessions, and reasoning traces.
The dataset construction and session format are designed to support reproducible analysis of agent behavior under long-horizon, tool-mediated interaction.

\subsection{Binary Corpus}

The binary corpus serves as a source of diverse analysis contexts, rather than the primary object of study.
We construct the binary corpus through a three-stage process.
First, we download firmware images from the Karonte dataset~\cite{redini2020karonte}, extract their filesystems, and collect 30,195 binaries from Linux-based firmware images.
Second, we apply Mango~\cite{gibbs2024operation}, a static analysis tool that identifies binaries likely to contain taint-style vulnerabilities (e.g., command injection and path traversal), reducing the corpus to 3,500 security-relevant binaries.
Third, we deduplicate binaries by filename and randomly select one binary
per unique filename, yielding 521 distinct binaries for analysis.
Filename-based deduplication is adopted because our focus is on \emph{token-level implicit patterns} rather than binary-specific vulnerability coverage.
Although binaries sharing the same filename may differ across compilation options or versions, they typically implement similar functionality.
This strategy avoids over-representation of popular binaries while preserving program diversity.

Table~\ref{tab:dataset} summarizes the corpus and session characteristics. The corpus spans multiple architectures, operating systems, and endianness; all binaries are ELF32 format, providing consistency while maintaining diversity in code structure and complexity.
These binaries contain known vulnerabilities and have been widely used as datasets in vulnerability detection research~\cite{redini2020karonte,feng2021snipuzz, wu2024your, chen2021sharing,gibbs2024operation}, which target taint-style vulnerabilities in firmware binaries.
Our goal is not exhaustive vulnerability coverage, but to observe sustained exploration behavior under realistic analytical conditions.

\begin{table}[!t]
    \centering
    \small
    \caption{Dataset and session overview.}
    \label{tab:dataset}
    \begin{tabular}{l r}
    \toprule
    Item & Count \\
    \midrule
        Total analysis binaries & 521 \\
        Architectures & ARM: 328; MIPS: 193 \\
        OS & Linux:519; Android:2 \\
        Endianness & Little Endian: 477; Big Endian: 44\\ 
    Format & ELF32: 521\\ \midrule
    Reasoning sessions & 521 \\
    Total reasoning steps & 99,563\\
    Median steps per session & 160.0 \\
    Average steps per session & 191.1 $\pm$ 154.9 \\
    \multicolumn{2}{l}{LLM:  \qquad DeepSeek-V3, GPT-5, Claude 3.5 Sonnet, Gemini 3.0} \\
    All sessions & 521$\times$4 \\
    \bottomrule
    \end{tabular}
\end{table}

\subsection{Session Generation}

Each binary is analyzed independently by an LLM agent equipped with a fixed tool interface.
Agents interact with binaries exclusively through standard program analysis tools (\texttt{radare2} and \texttt{Ghidra}, via the r2ghidra plugin) and receive no external guidance or human intervention.

For each binary and each LLM, we generate one continuous session—a temporally ordered trace of reasoning--action--observation steps—from initial reconnaissance to termination.
At each step, the trace records:
(1) the agent's reasoning state,
(2) the selected tool action, and
(3) the resulting observation returned by the tool.
The complete session format, initialization prompt, and JSONL record structure are specified in Appendix~\ref{app:session}.

Achieving long-horizon sessions (typically 130--300 steps) under LLM context window constraints requires explicit context management.
To this end, traces include \emph{LLM context reset} records, each representing a new LLM instance initialized with a compressed summary of prior analysis state.
The first context reset record contains the unified initialization prompt, while subsequent records preserve accumulated findings and task state.
Context resets allow traces to extend beyond single-context limits while preserving observable reasoning behavior; all tool actions and reasoning states remain directly recorded in the trace, enabling the study of long-horizon dynamics.
One reasoning segment within a larger vulnerability analysis session is illustrated in Appendix~\ref{app:trace-example}.

We analyze four LLMs (DeepSeek-V3, GPT-5, Claude~3.5~Sonnet, Gemini~3.0),
generating four independent sessions per binary.
For primary analysis, we select one representative session per binary to
avoid inter-session dependency.
Session selection prioritizes trace completeness and length and does not
filter based on vulnerability outcomes or analysis success, illustrated in Appendix~\ref{app:trace-generation}.
This setup produces traces where exploration behavior unfolds over time as a dynamic process.

\subsection{Trace Statistics}

Table~\ref{tab:dataset} summarizes the dataset characteristics. Sessions vary substantially in length, with a median of 160 steps and an average of 191 steps per session (standard deviation 154.9), reflecting the varying complexity of binary analysis tasks. The total dataset contains 99,563 reasoning steps, providing sufficient scale to identify stable token-level implicit patterns across diverse contexts.
The scale and variability of traces provide a basis for identifying stable patterns that are not tied to specific binaries or tasks. 
Note that the analyzed binaries predominantly contain real-world vulnerabilities: 38.0\% of analysis sessions (198 out of 521) identify at least one CWE-labeled vulnerability finding, with a total of 306 distinct vulnerability instances across the dataset (detailed statistics reported in Appendix~\ref{app:vuln-stats}). 
This ensures that observed behaviors arise from non-trivial security analysis tasks rather than trivial or empty exploration scenarios.

\section{Pattern}
\label{sec:pattern}

This section presents the four token-level patterns we extract from traces.


\subsection{Pattern Taxonomy}

We identify four recurring token-level patterns that characterize how LLM agents explore binary programs during vulnerability analysis. 
These patterns capture exploration-regulation behaviors emerging in traces under bounded reasoning capacity.
Pattern P4 (Knowledge-Guided Prioritization) corresponds to knowledge-guided prioritization behavior that ranks candidate paths based on prior knowledge and structural cues. 
Patterns P1–P3 describe observable dynamics in traces: early pruning behavior in early phases, path lock-in behavior during sustained analysis, and targeted backtracking behavior when revisiting deferred candidates.

Table~\ref{tab:pattern-summary} summarizes the core behaviors and primary benefits of these patterns. 
To extract patterns from traces, we adopt a rule-based approach that identifies observable structural features (e.g., revisits, path divergence, and persistence) from action–observation sequences.
The patterns are empirical regularities emerging in traces; the rules serve only as observable proxies for detection, not as definitions.
These patterns are not designed or programmed; they emerge consistently from token-level reasoning dynamics across sessions.
Full rule definitions and implementation details are provided in Appendix~\ref{app:pattern-detection} (Algorithms~\ref{alg:pattern1}–\ref{alg:pattern4}).

\begin{table}[!t]
  \small
  \centering
  \caption{Summary of token-level reasoning patterns in binary analysis.}
  \label{tab:pattern-summary}
  \begin{tabular}{p{1.8cm}|p{2.8cm}|p{2.8cm}}
  \hline
  \textbf{Pattern} & \textbf{Core Behavior} & \textbf{Primary Benefit} \\
  \hline
  
  P1: Early Pruning
  & Trace exhibits early pruning behavior: candidate paths discarded early, rarely revisited.
  & Efficient reduction of large search spaces.\\
  
  \hline
  
  P2: Path Lock-in
  & Trace exhibits path lock-in behavior: reasoning remains focused on a selected path.
  & Sustained contextual coherence during deep analysis.\\
  
  \hline
  
  P3: Targeted Backtracking
  & Trace exhibits targeted backtracking behavior: previously deferred paths revisited.
  & Recovery from incomplete or unproductive analysis paths.\\
  
  \hline
  
  P4: Knowledge-Guided Prioritization
  & Trace exhibits knowledge-guided prioritization behavior: paths ranked using prior knowledge.
  & Rapid prioritization of analysis directions.\\
  
  \hline
  \end{tabular}
  \end{table}

\subsection{Pattern 1: Early Pruning of Candidate Paths}

\begin{figure}[!t]
  \centering
  \includegraphics[width=0.85\linewidth]{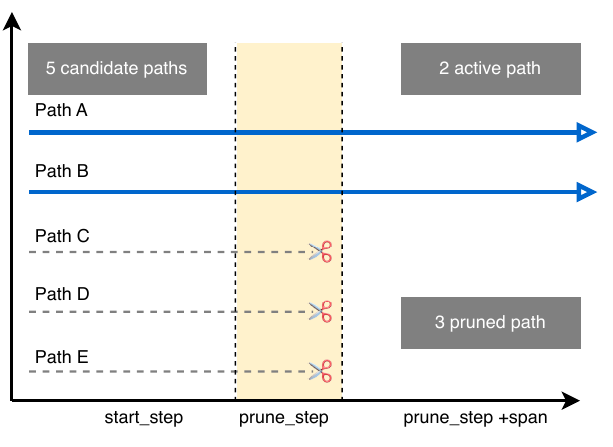}
  \caption{Pattern 1: pruning behavior in traces—candidate paths discarded early, rarely revisited.}
  \label{fig:pattern1}
\end{figure}

Binary vulnerability analysis often generates multiple candidate execution paths as functions, control-flow branches, and cross-references expand the exploration space.
Under bounded reasoning capacity, trace segments frequently exhibit early pruning behavior early in the analysis process, narrowing the candidate set. 
Instead of exploring alternatives uniformly, reasoning in the trace discards paths that appear unlikely to yield exploitable behavior. 
Once early pruning behavior occurs, subsequent reasoning steps rarely revisit the discarded paths, even when additional context later becomes available.
We label this pattern \textit{Early Pruning of Candidate Paths}. 
Figure~\ref{fig:pattern1} illustrates the typical temporal structure: an initial exploration phase generates multiple candidate paths, followed by early pruning behavior that removes several alternatives, after which reasoning continues primarily along the retained path.

\textbf{Detection Operation.}
Pattern~1 is identified when a trace segment exhibits the following observable sequence: 
(1) explicit reasoning about multiple candidate analysis paths, 
(2) a pruning decision that eliminates one or more alternatives based on diagnostic observations, and 
(3) continued reasoning that focuses exclusively on the retained path without revisiting discarded alternatives within a bounded observation window.

\textbf{Example Trace Snippet.}
In the \texttt{pptp-ondemand} binary session, the LLM agent initially explores several potential call sites: 
\textit{"The function is called from three locations in main. I need to examine each call site to understand the arguments passed, especially param\_2 and param\_3, to determine if they are tainted from external sources."} 
After several reasoning steps, the agent identifies a promising vulnerability path: 
\textit{"I have the full assembly of the function. I can see the buffer overflow clearly. At address 0x004011d8, there is 'addiu v0, fp, 0x118', which references the stack buffer..."} 
From this point onward, the analysis focuses exclusively on the identified path, while previously considered call sites are not revisited.

\subsection{Pattern 2: Path-Dependent Analysis Lock-in}

\begin{figure}[!t]
  \centering
  \includegraphics[width=0.85\linewidth]{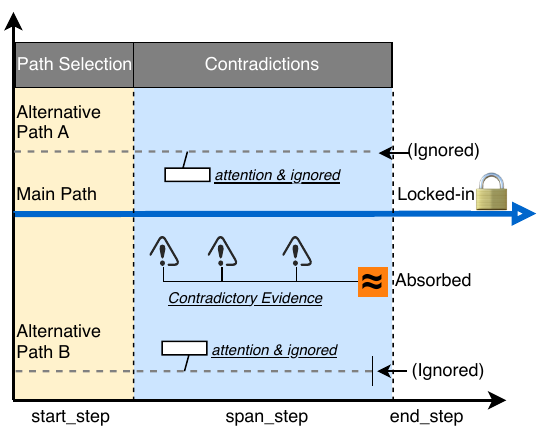}
  \caption{Pattern 2: lock-in behavior in traces—sustained reasoning within the same context, limited exploration of alternatives.}
  \label{fig:pattern2}
\end{figure}

Binary vulnerability analysis often requires extended reasoning along a single candidate execution path, such as tracing data propagation through multiple functions or examining a sequence of control-flow decisions.
However, switching analysis context between different candidate paths can disrupt reasoning coherence, particularly in long-horizon analyses.

Traces therefore often exhibit path lock-in behavior once an analysis path has been selected in the token stream.
Subsequent reasoning steps remain strongly conditioned on the established context, repeatedly examining related functions, instructions, and data flows associated with the chosen hypothesis.
Even when partial contradictions or ambiguous signals appear, the trace often continues within the same path rather than immediately switching to an alternative.
We label this pattern \textit{Path-Dependent Analysis Lock-in}.
Figure~\ref{fig:pattern2} illustrates the typical temporal structure: path-selection behavior early is followed by sustained reasoning within the same semantic context, while alternative paths receive little or no further exploration.

\textbf{Detection Operation.}
Pattern~2 is identified when a trace segment exhibits the following observable sequence:
(1) an explicit path-selection decision focusing analysis on a specific function, sink, or execution path,
(2) an extended reasoning span in which subsequent actions repeatedly operate within the same semantic context (e.g., examining related functions, addresses, or control-flow structures), and
(3) limited expansion of alternative paths despite occasional mentions.

\textbf{Example Trace Snippet.}
In the \texttt{3322ip} binary session, the trace exhibits path-selection behavior:
\textit{"The main function is large and complex. I can see several interesting things: it calls doSystemCmd at multiple locations (0xdf54, 0xe198, 0xe27c, 0xe628). The doSystemCmd is a dangerous sink (command injection). I need to identify taint sources that could reach doSystemCmd."}
Over the next several reasoning steps, path lock-in behavior emerges: the trace continues examining functions and instructions related to the \texttt{doSystemCmd} path, repeatedly tracing potential data flows toward this sink.
Even when intermediate observations weaken the initial hypothesis, the trace maintains focus on the same analysis path rather than switching to alternative sinks or unrelated functions.

\subsection{Pattern 3: Multi-Path Exploration with Targeted Backtracking}

\begin{figure}[!t]
  \centering
  \includegraphics[width=0.85\linewidth]{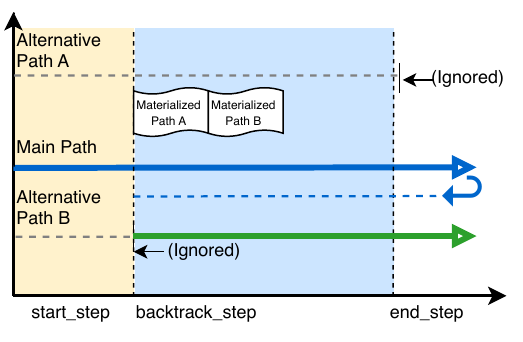}
  \caption{Pattern 3: backtracking behavior in traces—when the active path stalls, reasoning returns to a previously deferred candidate.}
  \label{fig:pattern3}
\end{figure}

Binary vulnerability analysis rarely follows a single linear reasoning path.
Early hypotheses may later prove incomplete or incorrect, requiring analysts to revisit alternative candidate paths that were previously considered but not fully explored.

Traces often exhibit multiple candidate paths as lightweight hypotheses within the reasoning context.
Only a subset of these paths is expanded at any given time, while others remain deferred but implicitly tracked as potential alternatives.
When the active analysis path reaches an impasse or when new evidence suggests that another path may be more promising, targeted backtracking behavior emerges—the trace returns to one of these previously deferred candidates and continues exploration.
We label this pattern \textit{Multi-Path Exploration with Targeted Backtracking}.
Figure~\ref{fig:pattern3} illustrates the typical temporal structure: multiple candidate paths are initially identified, one path is developed in depth, and a previously deferred candidate is later revisited and expanded.

\textbf{Detection Operation.}
Pattern~3 is identified when a trace segment exhibits the following sequence:
(1) multiple candidate paths are explicitly mentioned during early exploration,
(2) one candidate path is selected for deeper analysis while others remain undeveloped, and
(3) a previously deferred candidate path is later revisited and expanded after a substantial reasoning interval.

\textbf{Example Trace Snippet.}
In the \texttt{3322ip} binary session, the trace initially enumerates several potential analysis directions while inspecting program structure and function calls.
One path focusing on network-related functions is developed in depth.
Later in the trace, targeted backtracking behavior emerges: reasoning returns to a previously mentioned candidate involving the \texttt{doSystemCmd} sink:
\textit{"This function handles network connection setup and also calls doSystemCmd with format strings 'ip rule del to \%s' and 'ip rule add to \%s table wan \%d'..."}.
The trace then expands this deferred candidate path, combining information from earlier exploration with the newly examined function.

\subsection{Pattern 4: Knowledge-Guided Prioritization}

\begin{figure}[!t]
  \centering
  \includegraphics[width=0.85\linewidth]{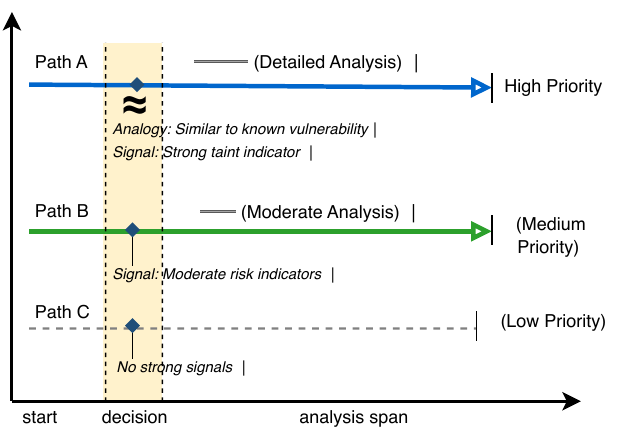}
  \caption{Pattern 4: prioritization behavior in traces—paths ranked early using prior knowledge, leading to uneven allocation of effort.}
  \label{fig:pattern4}
\end{figure}

Binary vulnerability analysis often exposes a large number of potential execution paths.
Fully exploring every candidate path is computationally infeasible, requiring analysts to prioritize which paths deserve deeper investigation.

Traces frequently exhibit knowledge-guided prioritization behavior, leveraging knowledge encoded in the model's pre-training.
Rather than waiting for complete information, reasoning in the trace evaluates candidate paths using partial evidence and structural cues, such as known dangerous functions, library calls, data-flow signals, or code patterns that resemble previously observed vulnerabilities.
Paths that appear more promising receive deeper exploration in subsequent steps, while others receive limited attention or remain deferred.
We label this pattern \textit{Knowledge-Guided Prioritization}.
Figure~\ref{fig:pattern4} illustrates the typical temporal structure: multiple candidate paths are identified, early reasoning exhibits knowledge-guided prioritization behavior assigning higher priority to some paths based on prior knowledge and partial signals, and subsequent analysis effort becomes unevenly distributed across these paths.

\textbf{Detection Operation.}
Pattern~4 is identified when a trace segment exhibits the following conditions:
(1) the agent explicitly ranks, selects, or prioritizes among multiple candidate paths, and
(2) the decision is justified using references to known vulnerability patterns, risky functions, structural cues, or partial data-flow evidence.
The prioritization decision and its justification must occur within the same reasoning step or in closely adjacent steps.

\textbf{Example Trace Snippet.}
In the \texttt{3322ip} binary session, the trace exhibits knowledge-guided prioritization behavior toward a command execution path:
\textit{"The main function calls doSystemCmd at multiple locations. The doSystemCmd function is a dangerous sink for command injection. I should identify possible taint sources that could reach this function."}
Based on the known risk associated with command execution sinks, the trace allocates detailed analysis effort to paths involving \texttt{doSystemCmd}, while other functions receive limited exploration.


\section{Empirical Analysis}
\label{sec:exp}

This section presents our empirical analysis of the reasoning patterns. 
We are attempting to answer three research questions.

\noindent\textbf{RQ1:} \textit{To what extent are the four patterns prevalent across analysis sessions, and how frequently are they invoked when present? } (\S~\ref{sec:exp:rq1})

\noindent\textbf{RQ2:} \textit{How do patterns transition and organize into sequential structures during reasoning?} (\S~\ref{sec:exp:rq2})

\noindent\textbf{RQ3:} \textit{How do different patterns show distinct behavioral characteristics as measured by trace-level metrics?} (\S~\ref{sec:exp:rq3})

\noindent 
Specifically, RQ1 evaluates whether patterns are consistently present across sessions; RQ2 examines whether patterns exhibit structured relationships rather than random occurrences; and RQ3 analyzes whether patterns correspond to distinct behavioral characteristics.

\textbf{Pattern Usage vs. Vulnerability Outcomes.}
While one may ask whether certain patterns lead to higher vulnerability discovery success, our study does not treat pattern usage as a performance predictor.
In our setting, pattern usage is primarily driven by binary complexity and reasoning dynamics, whereas vulnerability discovery outcomes depend on factors such as vulnerability presence, exploitability, and analysis difficulty.
As a result, there is no direct causal relationship between the frequency of specific patterns and vulnerability discovery outcomes.

\subsection{Pattern Prevalence and Density}
\label{sec:exp:rq1}

We analyze pattern prevalence (the proportion of sessions containing each pattern) and density (the frequency of pattern use when present). Table~\ref{tab:density} summarizes the key statistics.

\begin{table}[!t]
    \centering
    \small
    \caption{Pattern prevalence and density per session.}
    \label{tab:density}
    \begin{tabular}{l r r r r r r}
    \toprule
     & \makecell[r]{Sessions\\  w/ Pattern} & Coverage  & \makecell[r]{Total } & \makecell[r]{Avg / \\ Session} & \makecell[r]{Avg / Active \\  Session} & Max     \\
    \midrule
    P1      & 435 / 521           & 83.5\% & 3,339           & 6.41          & 7.68             & 44      \\
    P2      & 509 / 521           & 97.6\% & \makecell[r]{9,654}           & 18.53         & 18.99            & 77      \\
    P3      & 489 / 521           & 93.8\% & \makecell[r]{995}             & 1.91          & 2.03             & 5       \\
    P4      & 509 / 521           & 97.6\% & \makecell[r]{14,083}           & 27.03         & 27.71            & 278 \\
    \bottomrule
    \end{tabular}
\end{table}

\textbf{Pattern Prevalence}.
Table~\ref{tab:density} shows large differences in pattern prevalence: P2 and P4 appear in 97.6\% of sessions (509 out of 521), appearing in almost all reasoning sessions. P3 appears in 93.8\% of sessions (489 out of 521), while P1 appears in 83.5\% of sessions (435 out of 521). 
This variation shows that patterns are not uniformly present across all sessions. 
P1's lower prevalence (83.5\%) suggests it acts as a filtering mechanism—activated when needed for search control, but not mandatory for successful reasoning. 
P2 and P4 appear in almost all sessions (97.6\%), indicating they are fundamental components of LLM reasoning, appearing in almost every analysis session regardless of binary complexity or LLM model.

\textbf{Pattern Density per Session}.
We quantify pattern density as the number of pattern instances per session, measuring how often each reasoning behavior occurs during a single reasoning session.
Table~\ref{tab:density} shows large differences in usage frequency. 
P4 averages 27.7 instances per active session, far exceeding others. P2 averages 19.0 instances, P1 averages 7.7, while P3 averages only 2.0 instances per active session. These differences suggest that some patterns are used repeatedly (P4, P2), while others (P3) are used selectively and sparingly.
The maximum values show another important distinction: P4 shows extreme bursts (max=278), indicating that it can dominate decision-making in certain contexts. 
In contrast, P3 stays within a small range (max=5), never exceeding a small number of uses. This constraint suggests P3 acts as a limited recovery mechanism rather than a dominant mode of operation.


We further examine the overall patterns per session—the total number of pattern instances across all patterns within a session. This session-level view shows the intensity of pattern-driven reasoning across different analysis contexts. 
Table~\ref{tab:multi} shows the distribution of total pattern instances across sessions. Most sessions (36.08\%) contain 51-100 total pattern instances, with large differences ranging from sessions with very few instances (1-10) to sessions with extremely high activity (201+). This variation reflects differences in binary complexity, analysis depth, and reasoning requirements across different tasks.
This observation shows that the LLM agentic system uses a dynamic combination of complementary patterns rather than mutually exclusive modes. 
We observe similar pattern distributions across different architectures, LLM models, and compilation settings, further supporting their status as stable behavioral characteristics.

\begin{table}[t]
    \centering
    \small
    \caption{Distribution of total pattern instances per session.}
    \label{tab:multi}
    \begin{tabular}{c r r }
    \toprule
    Total Pattern Instances per Session & \# Sessions  & Percentage \\
    \midrule
0	&	6	&	1.15\% \\
1-10	&	41	&	7.87\% \\
11-30	&	104	&	19.96\% \\
31-50	&	135	&	25.91\% \\
51-100	&	188	&	36.08\% \\
101-200	&	40	&	7.68\% \\
201+	&	7	&	1.34\% \\
    \bottomrule
    \end{tabular}
\end{table}

\begin{figure}[!t]
    \centering
    \includegraphics[width=0.35\textwidth]{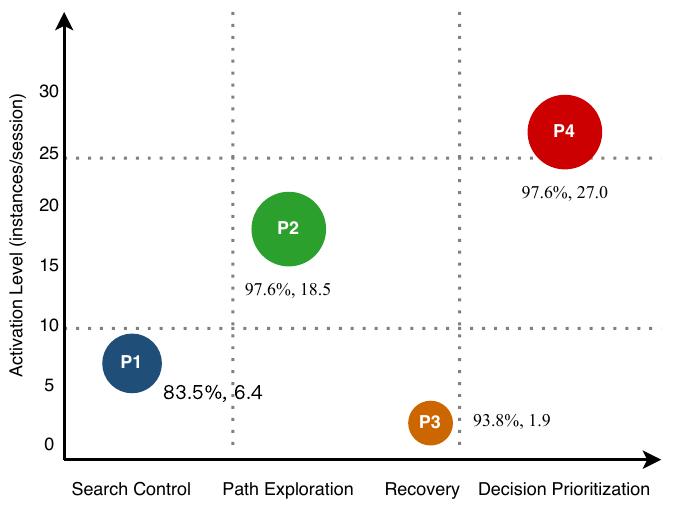}
    \caption{Pattern roles mapped along two dimensions: reasoning control (X-axis) and activation level measured by average instances per session (Y-axis).}
    \label{fig:pattern-role-map}
\end{figure}

\textbf{Pattern Roles}.
The density measurements reveal distinct operational characteristics that define each pattern's role in reasoning. Figure~\ref{fig:pattern-role-map} maps patterns along two dimensions: their role in reasoning control (X-axis) and activation characteristics (Y-axis), illustrating that patterns serve distinct roles rather than being redundant or arbitrary classifications.

\textit{P1 (Early Pruning)} appears in the region with moderate activation (83.5\% prevalence, 6.4 avg/session). Its selective coverage, combined with moderate frequency, suggests that it acts as a filtering mechanism—activated when needed to control search expansion but not required in every reasoning context. 
\textit{P2 (Path-dependent Lock-in)} occupies the region with high activation (97.6\% prevalence, 18.5 avg/session). Appearing in almost all sessions with high frequency, P2 forms the backbone of reasoning, providing the underlying framework that organizes path-dependent exploration throughout the session.
\textit{P3 (Multi-Path Exploration)} is positioned in the region with low activation (93.8\% prevalence, 1.9 avg/session). Despite appearing in most sessions, it has a very low frequency and stays within a small range (max=5). This combination—high presence but low usage—indicates P3 acts as a limited recovery mechanism: available when needed to correct reasoning failures, but used sparingly and with constraint, never dominating the reasoning process.
\textit{P4 (Knowledge-guided Prioritization)} dominates the region with the highest activation (97.6\% prevalence, 27.0 avg/session). With the highest frequency and extreme bursts (max=278), P4 dominates fine-grained decision-making, operating persistently throughout reasoning and showing bursts when multiple prioritization decisions cluster together. As the underlying decision-making mechanism that guides prioritization across reasoning stages, P4 provides the knowledge-guided foundation for decisions made in P1--P3.

\begin{figure}[!t]
    \centering
    \includegraphics[width=0.45\textwidth]{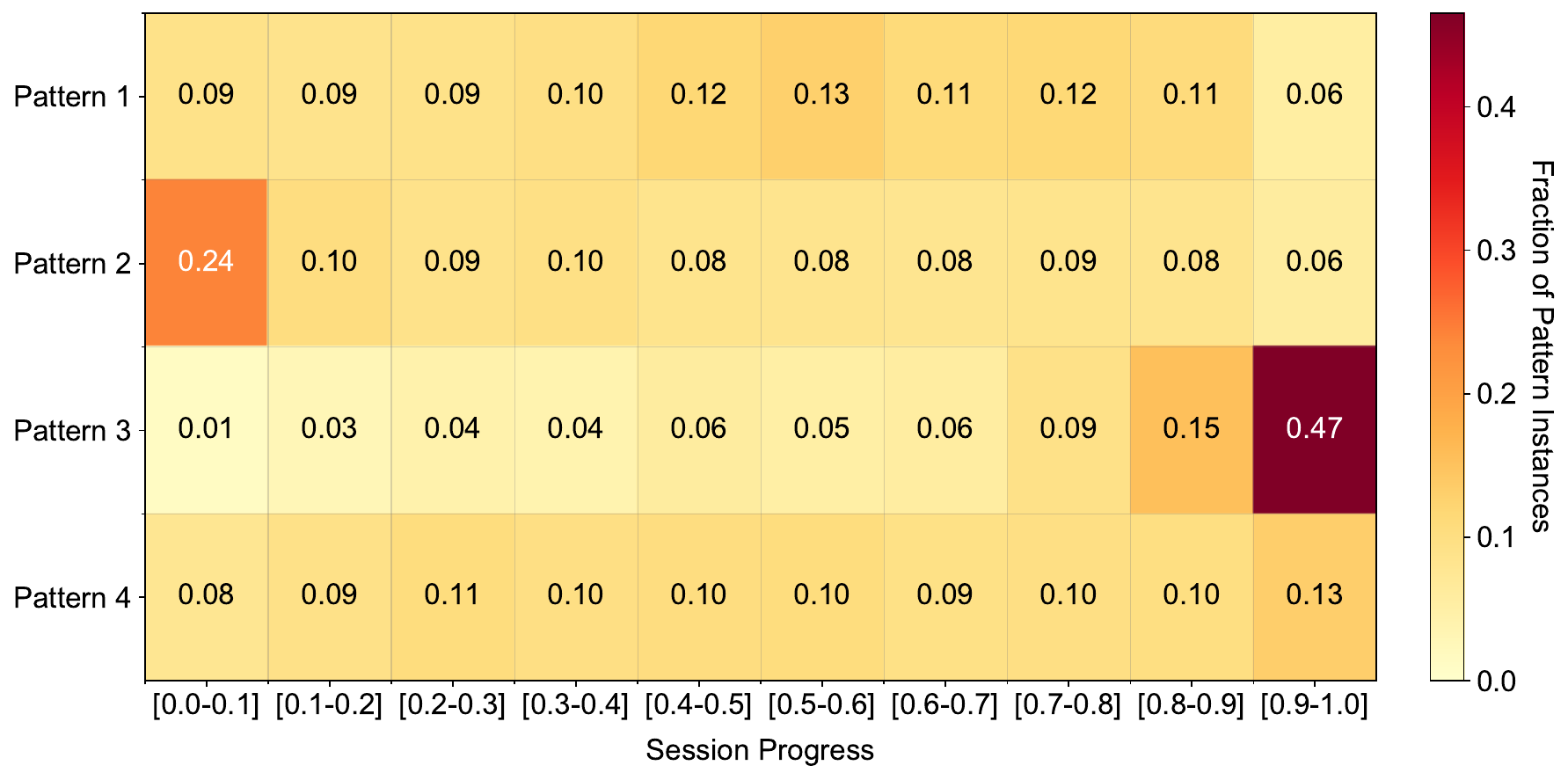}
    \caption{The heatmap showing the distribution of pattern instances across 10 phase bins (P1--P10) of normalized session progress. Each row represents one pattern, and each cell's color intensity encodes the fraction of pattern instances occurring in that phase. Rows are independently normalized to highlight phase bias rather than absolute frequency.}
    \label{fig:temporal-heatmap}
\end{figure}

\textbf{Temporal Coverage of Patterns}
We analyze when patterns occur during reasoning sessions by normalizing each session's timeline to [0,1] and partitioning it into 10 phase bins. Figure~\ref{fig:temporal-heatmap} shows the distribution of pattern instances across these phases, with each row normalized independently to highlight temporal bias.

The heatmap shows distinct temporal roles: \textit{Pattern 2 (Path Lock-in)} is early-biased (24.0\% in P1), reflecting initial hypothesis commitment. \textit{Pattern 1 (Early Pruning)} concentrates in the mid-phase (peak at P6, 12.9\%), as pruning decisions require accumulated contextual information. \textit{Pattern 3 (Multi-Path Exploration)} is late-biased (46.5\% in P10), consistent with its reactive role in correcting earlier decisions. \textit{Pattern 4 (Knowledge-guided Prioritization)} shows relatively uniform distribution (9--13\% across P3--P10), indicating opportunistic activation throughout the session. This phase-specific distribution shows that patterns serve distinct temporal roles rather than appearing randomly throughout reasoning.

\find{
    Takeaway 1: 
    LLM reasoning patterns are stable, repeatedly used components with distinct roles, not sporadic artifacts or configuration-specific behaviors.
}

\subsection{Pattern Sequential Structure and Inter-pattern Relationships}
\label{sec:exp:rq2}

Since patterns appear together within sessions, we examine how they organize into sequential structures through transitions and recurring sequences.

\begin{figure}[!t]
    \centering
    \begin{minipage}{0.23\textwidth}
        \centering
        \includegraphics[width=0.78\textwidth]{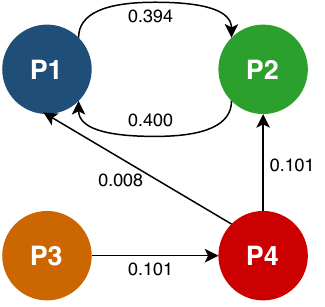}
        \caption{Pattern transition graph.}
        \label{fig:transition-graph}
    \end{minipage}
    \hfill
    \begin{minipage}{0.23\textwidth}
        \centering
        \small
        \begin{tabular}{l r r}
        \toprule
        Transition & Count & Prop. \\
        \midrule
        P2 → P1 & 1,947 & 0.400 \\
        P1 → P2 & 1,918 & 0.394 \\
        P3 → P4 & 492 & 0.101 \\
        P4 → P2 & 467 & 0.096 \\
        P4 → P1 & 40 & 0.008 \\
        P3 → P2 & 1 & 0.000 \\
        \bottomrule
        \end{tabular}
        \caption{Pattern transition frequency details.}
        \label{tab:transitions-block}
    \end{minipage}
\end{figure}

\textbf{Pattern Transitions}.
We analyze how patterns transition from one to another by collapsing consecutive identical patterns into blocks and examining block-level transitions. For example, P2, P2, P2, P1, P1, P4, P4, P2 becomes P2 → P1 → P4 → P2. 
Table~\ref{tab:transitions-block} and Figure~\ref{fig:transition-graph} show clear transition preferences. 
P2 and P1 form a bidirectional loop (P2→P1: 40.0\%; P1→P2: 39.4\%), together accounting for 79.4\% of all pattern switches, indicating they frequently alternate during reasoning. P3 transitions primarily to P4 (10.1\%), suggesting recovery actions often lead to prioritization decisions. P4 transitions to P2 (9.6\%) and P1 (0.8\%), indicating it frequently initiates commitment or pruning sequences. P3 rarely transitions to P2 (0.02\%), consistent with its targeted recovery role.

\begin{table}[!t]
    \centering
    \small
    \caption{Most frequent subsequences across all reasoning sessions.}
    \label{tab:subsequences}
    \begin{tabular}{l l r}
    \toprule
    Rank & Pattern Sequence & Frequency \\
    \midrule
    1 & P2 → P1 & 1,947 \\
    2 & P1 → P2 & 1,918 \\
    3 & P2 → P1 → P2 & 1,878 \\
    4 & P1 → P2 → P1 & 1,547 \\
    5 & P2 → P1 → P2 → P1 & 1,508 \\
    6 & P1 → P2 → P1 → P2 & 1,488 \\
    7 & P3 → P4 & 492 \\
    8 & P4 → P2 & 467 \\
    9 & P3 → P4 → P2 & 451 \\
    10 & P4 → P2 → P1 & 400 \\
    \bottomrule
    \end{tabular}
\end{table}

\textbf{Pattern Subsequences}.
We identify recurring sequences of length 2-4 extracted from block-level sequences to examine longer-range structure.
Table~\ref{tab:subsequences} shows two major reasoning macro-structures. 
First, P2 and P1 form alternating loops: P2→P1 (1,947) and P1→P2 (1,918) are the most frequent length-2 sequences, while extended alternations P2→P1→P2 (1,878) and P1→P2→P1 (1,547) dominate length-3 sequences. Length-4 sequences continue this pattern, with P2→P1→P2→P1 (1,508) and P1→P2→P1→P2 (1,488) being most common. This persistent alternation suggests a core reasoning routine that repeatedly switches between path lock-in (P2) and selective pruning (P1).
Second, P3 and P4 form recovery-to-prioritization sequences: P3→P4 (492) and P3→P4→P2 (451) indicate that recovery actions (P3) frequently lead to prioritization decisions (P4), which then transition to path lock-in (P2). These recurring sequences show that patterns combine into predictable macro-structures rather than appearing randomly.

\begin{figure}[!t]
    \centering
    \includegraphics[width=0.35\textwidth]{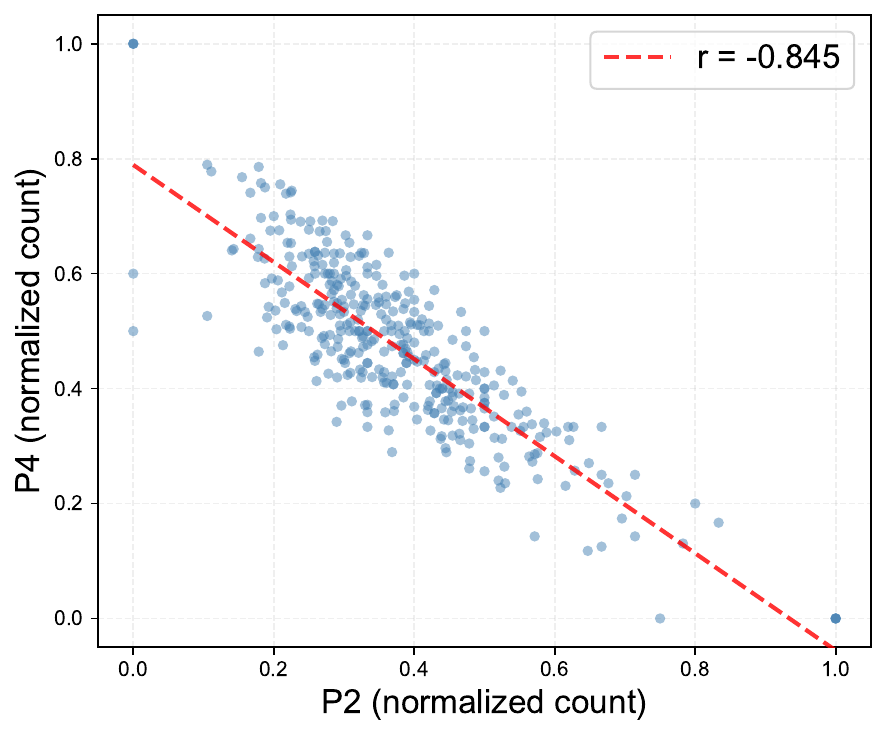}
    \caption{Joint distribution of normalized pattern counts: P2 vs P4. Each point represents a session. The strong negative correlation (r = -0.845) indicates complementary resource allocation: high P2 corresponds to low P4, and vice versa.}
    \label{fig:load-sharing}
\end{figure}

\textbf{Pattern Coordination}.
We examine whether patterns show systematic relationships in their usage intensity across sessions. We normalize pattern counts per session and analyze correlations between normalized counts.
Figure~\ref{fig:load-sharing} shows the joint distribution of normalized pattern counts: P2 vs P4. 
P2 and P4 show a strong negative correlation (r = -0.845): when P2 is above its median normalized count, P4 averages 0.377; when P2 is below its median, P4 averages 0.566. This indicates that path lock-in (P2) and decision prioritization (P4) function as complementary mechanisms. As the underlying decision-making mechanism, P4's prioritization decisions inform when paths are committed to (P2), pruned (P1), or reconsidered (P3), creating systematic relationships in pattern usage. 
P1 shows moderate negative correlations with other patterns (r = -0.216 to -0.246), suggesting it acts as a selective mechanism. P3 shows weak correlations (r = -0.075 to -0.126), consistent with its targeted recovery role that operates independently of the main reasoning flow.
Together, these analyses show that patterns organize into structured sequences with clear transition preferences and recurring macro-structures, rather than appearing randomly or independently during reasoning.

\find{
    Takeaway 2:
    Reasoning patterns operate as a coordinated system with predictable structural relationships, rather than independent or interchangeable behaviors.
}

\subsection{Pattern Behavioral Characteristics}
\label{sec:exp:rq3}

We quantify how different reasoning patterns show concrete action structures by measuring behavioral metrics over pattern-aligned trace segments.
Importantly, these metrics are computed independently of the pattern extraction rules, providing an external validation of pattern distinctiveness.
For each pattern instance, metrics are computed over the corresponding contiguous trace segment (see Appendix~\ref{app:metric} for computation details).

\begin{table}[t]
    \centering
    \small
    \caption{Metrics statistics: mean, median, and standard deviation are reported for each metric across all pattern instances.}
    \label{tab:pattern-metrics}
    \begin{tabular}{l l r r r }
    \toprule
    \textbf{Pattern} & \textbf{Metric} & \textbf{Mean} & \textbf{Median} & \textbf{Std} \\
    \midrule
    \multirow{4}{*}{P1 (3,339)} & Path Length ($L$) & 19.9 & 9.0 & 31.5  \\
    & Branching Factor ($B$) & 1.035 & 1.000 & 0.141   \\
    & Forward Step Ratio ($F$) & 0.869 & 0.947 & 0.226   \\
    & Backtrack Count ($R$) & 0.175 & 0.000 & 0.534   \\
    \midrule
    \multirow{4}{*}{P2 (9,654)} & Path Length ($L$) & 8.9 & 7.0 & 6.3   \\
    & Branching Factor ($B$) & 1.051 & 1.000 & 0.197   \\
    & Forward Step Ratio ($F$) & 0.944 & 1.000 & 0.097   \\
    & Backtrack Count ($R$) & 0.039 & 0.000 & 0.206   \\
    \midrule
    \multirow{4}{*}{P3 (995)} & Path Length ($L$) & 5.3 & 3.0 & 6.3   \\
    & Branching Factor ($B$) & 1.040 & 1.000 & 0.187   \\
    & Forward Step Ratio ($F$) & 0.860 & 1.000 & 0.251   \\
    & Backtrack Count ($R$) & 1.000 & 1.000 & 0.000   \\
    \midrule
    \multirow{4}{*}{P4 (14,083)} & Path Length ($L$) & 6.0 & 6.0 & 0.6   \\
    & Branching Factor ($B$) & 1.063 & 1.000 & 0.211   \\
    & Forward Step Ratio ($F$) & 0.878 & 0.857 & 0.147   \\
    & Backtrack Count ($R$) & 0.060 & 0.000 & 0.250   \\
    \bottomrule
    \end{tabular}
\end{table}

\textbf{Pattern-conditioned Action Metrics}.
We measure four metrics to quantify behavioral differences across patterns: path length ($L$), branching factor ($B$), forward step ratio ($F$), and backtrack count ($R$). 
Table~\ref{tab:pattern-metrics} reports mean, median, and standard deviation for each metric across all instances of each pattern. 

\textit{Early Pruning (P1)} shows the longest and most variable paths (mean $L=19.9$), with low branching (median $B=1.0$), high forward commitment (median $F=0.947$), and minimal backtracking. This indicates sustained linear progression with early elimination of candidate paths.
\textit{Path Lock-in (P2)} is characterized by short, stable paths (median $L=7.0$), the strongest forward commitment (median $F=1.0$), and almost no backtracking. These properties reflect deterministic, path-dependent progression where early decisions constrain subsequent analysis.
\textit{Multi-Path Exploration (P3)} shows short paths (median $L=3.0$) with consistent backtracking (median $R=1.0$), confirming its targeted recovery role.
\textit{Knowledge-guided Prioritization (P4)} shows highly stable path lengths (mean and median $L=6.0$), moderate forward progression, and minimal backtracking. Compared to P2, P4 shows slightly higher branching and lower forward ratios, suggesting flexible prioritization guided by analogical cues.

\begin{table}[t]
    \centering
    \small
    \caption{Exploration vs. Path Selection metrics. Branching Factor ($B$) measures exploration breadth, while Pruning Rate ($P$) measures path elimination aggressiveness.}
    \label{tab:exploration-commitment}
    \begin{tabular}{l r r r r}
    \toprule
    \textbf{Pattern} & &  & \textbf{Exploration} & \textbf{Path Selection} \\
    & \textbf{$B$: Mean} & \textbf{$P$: Mean} & \textbf{Level} & \textbf{Level} \\
    \midrule
    P1 & 1.035  & 0.118  & Low & High \\
    P2 & 1.051  & 0.043& Medium & Low \\
    P3 & 1.040 & 0.064  & Medium & Medium \\
    P4 & 1.063  & 0.078  & High & Medium \\
    \bottomrule
    \end{tabular}
\end{table}

\textbf{Exploration vs. Path Selection}.
We analyze how different patterns balance exploration breadth and path selection certainty using branching factor ($B$) and pruning rate ($P$). Branching factor measures how many options are considered at decision points, while pruning rate measures how aggressively candidate paths are eliminated. Together, these metrics show how patterns trade off between exploring multiple possibilities versus narrowing down to specific paths.

Table~\ref{tab:exploration-commitment} shows mean and median values of $B$ and $P$ for each pattern.
P1 shows the lowest branching factor ($B=1.035$) and the highest pruning rate ($P=0.118$), indicating limited exploration with aggressive path elimination. This reflects early narrowing of the search space.
P2 shows a slightly higher branching factor but the lowest pruning rate ($P=0.043$). Rather than actively eliminating paths, this pattern reflects a state where the reasoning path has already stabilized, reducing the need for further pruning. Exploration is moderate, but path selection emerges implicitly through persistence rather than explicit elimination.
P3 shows intermediate characteristics. Both branching factor and pruning rate remain moderate, reflecting a balance between maintaining multiple candidate paths and selectively discarding them. This preserves flexibility while enabling controlled rollback when necessary.
P4 shows the highest branching factor ($B=1.063$) with a moderate pruning rate. This suggests knowledge-guided decisions consider a broader set of alternatives before selection, leading to higher exploratory breadth while still narrowing down to specific paths.

\find{
    Takeaway 3:
    Patterns show distinct exploration–path selection characteristics: P1 emphasizes rapid path narrowing, P2 reflects implicit selection via persistence, P3 maintains a balanced state, and P4 favors broader exploration before selection.
}

\textbf{Pattern-conditioned Tool Usage Topology}.
Beyond path-level behavior, reasoning patterns also appear in the structure of tool usage. Different patterns correspond to distinct command usage patterns, reflecting how the LLM organizes its interaction with analysis tools during reasoning. We analyze tool usage topology by measuring six metrics over pattern-aligned trace segments: command diversity (number of distinct command types used), sequence length (total tool invocations), max depth (maximum consecutive repetition of the same command), max fan-out (maximum number of different commands following a given command), cycle presence (whether command transitions form cycles), and transition entropy (randomness of command transitions).
To illustrate the internal structure of a reasoning segment, Appendix~\ref{app:trace} presents an annotated 26-step tool interaction extracted from a longer session (187 steps) that ultimately identified a command injection vulnerability in a binary file.

Table~\ref{tab:tool-topology} reports these metrics for each pattern. Early Pruning (P1) shows the highest command diversity (5.96) and longest sequences (17.48), consistent with broad exploration requiring multiple tool types. Path Lock-in (P2) shows the highest cycle rate (92.5\%), indicating repetitive command patterns that reflect stable, committed reasoning trajectories. Multi-Path Exploration (P3) has the lowest diversity (2.68) and lowest transition entropy (0.23), showing deterministic tool usage during targeted recovery. Knowledge-guided Prioritization (P4) shows moderate diversity (3.29) with a high cycle rate (82.6\%), suggesting structured tool usage patterns that support semantic comparisons.
These results show that reasoning patterns correspond to distinct tool usage graph structures. High diversity and long sequences in P1 reflect exploration-phase tool usage; high cycle rates in P2 and P4 indicate repetitive command patterns supporting committed or structured reasoning; low diversity and entropy in P3 show deterministic tool selection during recovery. Together, these topological differences provide additional evidence that patterns capture systematic organizational principles in LLM reasoning.

\begin{table}[t]
    \centering
    \small
    \caption{Tool usage topology metrics by pattern type.}
    \label{tab:tool-topology}
    \begin{tabular}{l r r r r r r}
    \toprule
     & \textbf{Diversity} & \textbf{Length} & \textbf{Depth} & \textbf{Fan-out} & \textbf{Cycles} & \textbf{Entropy} \\
    & \textbf{Mean} & \textbf{Mean} & \textbf{Mean} & \textbf{Mean} & \textbf{(\%)} & \textbf{Mean} \\
    \midrule
    P1 & 5.96 & 17.48 & 2.77 & 2.65 & 70.9 & 0.60 \\
    P2 & 4.49 & 8.17 & 2.47 & 2.10 & 92.5 & 0.51 \\
    P3 & 2.68 & 4.71 & 1.76 & 1.22 & 51.4 & 0.23 \\
    P4 & 3.29 & 4.99 & 2.07 & 1.59 & 82.6 & 0.35 \\
    \bottomrule
    \end{tabular}
\end{table}

\find{Takeaway 4: 
Reasoning patterns show distinct tool usage structures, indicating systematic differences in how LLMs organize tool interactions.
}

\section{Discussion and Implications}
\label{sec:discuss}


\textbf{Implicit Control from Token-Level Reasoning.}
A central finding is that exploration control in LLM-based binary analysis is not explicitly programmed, but emerges implicitly from token-level sequential reasoning.
Unlike traditional systems where prioritization, pruning, and backtracking are implemented as explicit data structures and algorithms, LLM agents exhibit these behaviors without maintaining any explicit representation of the search state.
The four patterns demonstrate that key control functions—path selection, commitment, revision, and prioritization—can arise directly from the token generation process.
Early pruning emerges because LLMs cannot maintain all candidate paths simultaneously; path lock-in occurs because token-level reasoning creates strong contextual dependencies; backtracking reflects recovery from path failures; prioritization leverages pre-trained semantic knowledge to guide decisions.
This shifts the perspective of agent design: instead of specifying control logic, systems may rely on and shape the model's implicit reasoning dynamics.

\textbf{Semantic-Guided vs.\ Structural Exploration.}
LLM-driven exploration differs fundamentally from prior methods. Depth-first and breadth-first search require explicit graph structures with deterministic traversal; taint analysis operates with precise data-flow tracking. In contrast, LLM traversal is \emph{semantic-guided}: decisions are driven by learned semantic associations rather than structural reachability. When encountering a function call or branch, the LLM interprets semantic meaning and uses pre-trained knowledge to infer relevance to vulnerability discovery, enabling prioritization of semantically promising paths over structurally similar but semantically irrelevant ones.

\textbf{From One-Pass Pipelines to Iterative Analysis.}
Our results highlight a fundamental shift in how static analysis participates in vulnerability discovery. Traditional approaches follow a one-pass paradigm, where analysis constructs a global representation and reasoning operates over this fixed structure. LLM-based systems operate iteratively: analysis and reasoning are interleaved across hundreds of steps, with decisions continuously revised based on partial observations. The identified patterns reveal how this iterative process is structured—early pruning regulates search space growth, path lock-in sustains deep analysis, backtracking enables recovery, and prioritization guides exploration under uncertainty.

\textbf{Patterns as Structural Components, Not Heuristics.}
The four patterns should not be interpreted as task-specific heuristics or incidental behaviors. Their stability across binaries, distinct temporal roles, and structured transition relationships (e.g., the bidirectional loop between lock-in and pruning) indicate that they function as fundamental organizing principles for long-horizon reasoning. This perspective elevates patterns from descriptive observations to structural components that enable tractable exploration under bounded reasoning capacity.

\section{Threats to Validity}

\textbf{Task, System, and Model Dependence.}
Our analysis is conducted on reasoning traces generated during vulnerability-oriented binary analysis tasks, under a fixed system design and prompting protocol with a specific LLM. 
Task objectives (e.g., vulnerability discovery), system orchestration (e.g., iterative analysis), and model choice may influence the frequency or dominance of certain behaviors. 
However, the identified patterns are derived from observable decision-level transitions in long-horizon traces, rather than from task-specific heuristics or architecture-dependent features. 
We expect these behaviors to generalize to other security analysis tasks that require sequential evidence accumulation and adaptive exploration.

\textbf{Behavioral Interpretation and Scope.}
The identified patterns are inferred from externally observable reasoning traces, without claims about internal model states or cognitive mechanisms. 
Our analysis adopts a pattern-based perspective, focusing on the token-level implicit patterns that govern commitments, revisions, and path-selection decisions that directly affect analysis outcomes. 
This work prioritizes systematic pattern identification over exhaustive quantification; while grounded in concrete trace evidence, we do not claim complete coverage of all possible behaviors or configurations. 
Instead, the goal is to establish a trace-level foundation that enables future benchmarking, measurement, and controlled comparison across models and tools.

\section{Related Work}
\label{sec:related}



\textbf{A. Symbolic Execution and Static Analysis.}
Symbolic execution and static analysis techniques address challenges such as path explosion, state pruning, and scalability~\cite{stephens2016driller,yun2018qsym,poeplau2020symbolic,vadayath2022arbiter,gao2024faster,zhao2024leveraging,gibbs2024operation}. 
These approaches typically operate within a fixed analysis pipeline, where exploration strategies are explicitly defined. 
Prior work focuses on improving efficiency, coverage, and scalability of these mechanisms. 
In contrast, our work studies how token-level exploration behaviors arise without explicit implementation, emerging instead from token-level reasoning in LLM-driven analysis.

\textbf{B. Automated Vulnerability Discovery.}
Automated vulnerability discovery systems, including fuzzing and learning-based approaches, explore large search spaces using input mutation, statistical sampling, or heuristic guidance~\cite{xiao2025housefuzz,liu2025detecting,scharnowski2022fuzzware,zheng2019firm,gotovchits2018saluki,qasem2024octopustaint,feng2021snipuzz,li2022muafl}. 
While they incorporate implicit or explicit prioritization strategies, they do not expose or analyze the step-by-step decision processes that govern exploration. 
Our work instead focuses on characterizing the structure of exploration behavior itself, independent of specific optimization objectives.



\textbf{C. LLM-based Program Analysis and Agentic Systems.}
LLMs have been applied to program analysis and binary security tasks, including vulnerability detection, code reasoning, and tool-assisted exploration~\cite{feng2023prompting,li2023hitchhiker,pearce2023examining,deng2023large,lemieux2023codamosa,wang2024sanitizing,liu2025llm,jifirmagent,abramovich2025enigma}. 
Recent agentic systems further extend LLM capabilities to multi-step workflows, such as code repair and issue resolution, through tool use and iterative interaction~\cite{yang2024swe,zhang2024autocoderover,xia2024agentless,miserendino2025swelancer,chen2025locagent,li2025chainofagents}. 
These systems demonstrate the effectiveness of LLMs in long-horizon tasks, but primarily focus on task performance, prompting strategies, or system design. 
The internal reasoning process—particularly how exploration decisions evolve over long traces—remains unexamined. 
In contrast, our work provides a trace-level analysis of LLM-driven exploration, treating long-horizon reasoning behavior as the primary object of study.

\section{Conclusion}
\label{sec:conclusion}

This paper presents the first trace-level study of long-horizon LLM-driven binary analysis, showing that multi-pass reasoning gives rise to structured, token-level implicit patterns that govern exploration behavior.
Across 521 binaries and 99,563 reasoning steps, we identify four dominant patterns—early pruning, path-dependent lock-in, targeted backtracking, and knowledge-guided prioritization—that emerge directly from token-level sequential reasoning rather than explicit program logic.

Our findings suggest a shift in perspective: in LLM-based systems, exploration control is not implemented as explicit algorithms, but arises as an emergent property of token-level reasoning.
The identified patterns constitute a minimal set of structural mechanisms that organize long-horizon analysis under bounded context and uncertainty, providing an abstraction of how LLM agents regulate exploration over time.
This perspective opens new directions for both system design and analysis. Future work may build controllable and reliable LLM-based analysis systems by explicitly supporting these implicit mechanisms, as well as extend trace-level methodologies to study long-horizon reasoning in other domains beyond binary security.

\bibliographystyle{ACM-Reference-Format}
\bibliography{ref/ref_online, ref/ref_security, ref/ref_firmware, ref/ref_llm, ref/ref_code, ref/other-ref}

\appendix

\section{Session Generation Details}
\label{app:session}

\subsection{Initialization Prompt}

All sessions in our dataset use the same initialization prompt, which establishes the task and analysis requirements. 
The prompt serves solely as a task specification and does not encode system-specific constraints or implementation details.

The unified initialization prompt used across all sessions:
\begin{lstlisting}[breaklines=true,breakatwhitespace=false,basicstyle=\ttfamily\small,frame=single,escapeinside={(*@}{@*)},keepspaces=true,columns=fullflexible]
Please analyze the binary file comprehensively based on user 
core requirements. Current file being analyzed is: (*@\textcolor{blue}{[binary\_name]}@*). 
User core requirements are: Comprehensively analyze the 
specified binary file, with the core goal of identifying 
feasible exploitation chains from untrusted input sources 
(such as network, environment variables, IPC, etc.) to 
final dangerous operations. Always focus on precise analysis 
of current specific tasks, avoiding false negatives and 
false positives.

(*@\textcolor{blue}{\#\#\# Requirements}@*):
- Follow the function call protocol strictly
- Provide detailed analysis for each finding
- Use precise addresses and instruction locations
\end{lstlisting}

\noindent This prompt defines the analysis objective (vulnerability discovery through source-to-sink path identification) but does not prescribe specific reasoning strategies or exploration patterns; these emerge from the LLM's interaction with the binary under the given task constraints.

\subsection{Complete Session Format Example}

Each session is stored as a JSONL file, where each line represents a single record in the trace. 
The session is a sequence of reasoning-action-observation cycles. Below shows two representative record types:

\paragraph{Assistant record (thought and action):}
This record type captures the agent's internal reasoning state (\texttt{thought}) and selected tool action (\texttt{action}, \texttt{command}, \texttt{status}). The \texttt{thought} field contains the reasoning process, while \texttt{action} and \texttt{command} specify what tool action to take.

\begin{lstlisting}[breaklines=true,breakatwhitespace=false,basicstyle=\ttfamily\small,frame=single]
{
  "role": "assistant",
  "thought": "I've seen the main function. It opens /dev/ac0, uses getopt to parse command line arguments. The taint sources are command line arguments (argv). The program seems to be a utility to set/get AC entries via ioctl to a device. There are functions sym.SetAcEntry and sym.GetAcEntry that likely perform ioctl calls. I need to analyze those functions to see if there are any vulnerabilities like buffer overflows due to strcpy or other unsafe operations.",
  "action": "r2",
  "command": "pdf @sym.SetAcEntry",
  "status": "continue"
}
\end{lstlisting}

\paragraph{Tool record (observation):}
This record type captures the tool execution result (\texttt{type}, \texttt{tool}, \texttt{command}, \texttt{result}). The \texttt{result} field contains the observation returned by the tool, which is the partial program information that the agent uses to update its reasoning state.

\begin{lstlisting}[breaklines=true,breakatwhitespace=false,basicstyle=\ttfamily\small,frame=single]
{
  "role": "tool",
  "type": "tool_result",
  "tool": "r2",
  "command": "pdf @sym.SetAcEntry",
  "result": " ; CALL XREFS from main @ 0x4013cc(r)\n[320: sym.SetAcEntry (int32_t arg1, int32_t arg2, ...);\n| 0x00401540      d8ffbd27       addiu sp, sp, -0x28\n| 0x00401544      2400bfaf       sw ra, (var_24h)\n| 0x004015e0      8080828f       lw v0, -sym.imp.ioctl(gp)\n| 0x004015ec      09f82003       jalr t9\n| 0x0040167c      00000000       nop"
}
\end{lstlisting}

\begin{table*}[t]
   \centering
   \small
   \caption{Annotated radare2 command trace (26 steps). Each command corresponds to a localized analysis operation that reveals a specific program property.}
   \label{tab:annotated-trace}
   \begin{tabular}{r l l l}
   \toprule
   \textbf{Step} & \textbf{Command} & \textbf{Immediate Purpose} & \textbf{Context Gained / Used} \\
   \midrule
   0  & \texttt{i} 
      & Initialize binary metadata 
      & Architecture, bitness, entry points \\
   1  & \texttt{afl} 
      & Recover function list 
      & Function boundaries and symbols \\
   2  & \texttt{is\textasciitilde main} 
      & Locate main-related symbols 
      & Candidate entry function \\
   3  & \texttt{pdf @ entry0} 
      & Inspect entry function 
      & Control-flow of program start \\
   4  & \texttt{pdf @ fcn.0000db2c} 
      & Expand to adjacent function 
      & Callee function structure \\
   5  & \texttt{pdf @ fcn.0000b208} 
      & Continue function-space exploration 
      & Alternative execution path \\
   6  & \texttt{pdf @ fcn.0000ac9c} 
      & Scan related function 
      & Broader function context \\
   7  & \texttt{iz} 
      & Enumerate strings 
      & Candidate semantic anchors \\
   8  & \texttt{axt @@ sym.imp.doSystemCmd} 
      & Resolve cross-references 
      & Call sites invoking system command \\
   9  & \texttt{pd 20 @ 0xdf54} 
      & Inspect instructions near call site 
      & Local instruction semantics \\
   10 & \texttt{pd -30 @ 0xdf54} 
      & Backward inspection 
      & Preceding control dependencies \\
   11 & \texttt{axt @@ reloc.wanid} 
      & Resolve additional data references 
      & Alternative semantic anchor \\
   12 & \texttt{pdf @ fcn.0000b514} 
      & Jump to referenced function 
      & New function-level context \\
   13 & \texttt{is\textasciitilde doSystemCmd} 
      & Validate symbol relevance 
      & Confirmation of target semantics \\
   14 & \texttt{axt @@ reloc.ifname} 
      & Explore related references 
      & Configuration-dependent behavior \\
   15 & \texttt{pdg @ fcn.0000b208} 
      & High-level decompilation 
      & Abstract control/data flow \\
   16 & \texttt{pd 10 @ 0xb428} 
      & Instruction-level refinement 
      & Concrete execution details \\
   17 & \texttt{/a mov r0, 0xc} 
      & Pattern search 
      & Potential argument setup \\
   18 & \texttt{pd 5 @ 0xb458} 
      & Verify matched location 
      & Local confirmation \\
   19 & \texttt{pd -10 @ 0xb458} 
      & Context expansion 
      & Backward dependency check \\
   20 & \texttt{pd 10 @ 0xe198} 
      & Inspect another site 
      & Comparative behavior \\
   21 & \texttt{pd -20 @ 0xe198} 
      & Backtracking 
      & Control-flow consistency \\
   22 & \texttt{pd -10 @ 0xe27c} 
      & Further backward tracing 
      & Execution preconditions \\
   23 & \texttt{axt @@ sym.imp.GetIniFileValue} 
      & Trace configuration usage 
      & External input dependency \\
   24 & \texttt{pd 20 @ 0x9d58} 
      & Final local inspection 
      & Confirmation of hypothesis \\
   25 & \texttt{finish} 
      & Terminate session 
      & --- \\
   \bottomrule
   \end{tabular}
\end{table*}

\noindent Key fields:
\begin{itemize}
\item \texttt{role}: Identifies the record type (LLM context reset, assistant, tool)
\item \texttt{thought}: Reasoning state inferred from assistant records
\item \texttt{action}: Type of action taken (tool\_invocation, etc.)
\item \texttt{command}: Specific tool command issued
\item \texttt{result}: Tool output or observation returned
\item \texttt{status}: Execution status (executed, failed, etc.)
\end{itemize}

\noindent This format enables reconstruction of the complete reasoning-action-observation cycle for each session. The initialization prompt in the first LLM context reset record establishes the session context, while subsequent context reset records maintain continuity by preserving accumulated analysis state. This structure supports pattern detection and behavioral analysis while enabling long-horizon session continuity.

\section{Annotated Tool-Interaction Trace}
\label{app:trace}

\subsection{Trace Generation}
\label{app:trace-generation}

Achieving long-horizon analysis sessions requires maintaining reasoning continuity across extended interactions.
Because LLMs operate under finite context window constraints, we employ periodic \emph{context reset} operations during a session.
A context reset instantiates a new LLM instance with a fresh context window.
The initial context includes the task description, available tools, and analysis objectives.
Subsequent resets include a compact summary of accumulated analysis state, including key findings and the current exploration focus.
These summaries are generated automatically and do not introduce new information beyond what is already present in the trace.
Context resets serve only to preserve session continuity and do not alter the agent’s action space or decision logic.
All reasoning steps, tool actions, and observations remain explicitly recorded in the trace.
Our behavioral analysis operates solely on these observable elements and is agnostic to the reset boundaries.

As mentioned, we analyze four LLMs (DeepSeek-V3, GPT-5, Claude~3.5~Sonnet, Gemini~3.0),
generating four independent sessions per binary.
Session selection prioritizes trace completeness and quality. We discard incomplete or failed sessions, including runs that terminate prematurely due to tool errors or fail to progress beyond initial reconnaissance. If exactly one model produces a complete trace (reaching a minimum length threshold of 130 steps and terminating normally), we select that session. If multiple models produce complete traces, we select the longest trace. If all traces are incomplete or of similar length, we select the trace with the median length. We do not filter traces based on vulnerability outcomes, coverage, or success, as our analysis focuses on behavioral structure rather than effectiveness alone.

\subsection{Trace Example}  
\label{app:trace-example}

This illustrative trace corresponds to a single analysis session consisting of 26 radare2 commands, as shown in Table~\ref{tab:annotated-trace}.
The trace contains 10 distinct command types, with instruction-level disassembly commands (\texttt{pd}) dominating the interaction (9/26), followed by function-level disassembly (\texttt{pdf}, 5/26) and cross-reference queries (\texttt{axt}, 4/26). The trace exhibits multiple abstraction shifts between program-level, function-level, and instruction-level analysis, as well as repeated bidirectional inspection around the same addresses.

\textbf{Trace Logic Summary.}
This trace follows a repeated zoom-in/zoom-out reasoning structure. The agent first establishes a global structural view (Steps 0–3), then explores the function space (Steps 4–6), introduces semantic anchors via strings and cross-references (Steps 7–8), and iteratively validates hypotheses through localized bidirectional disassembly (Steps 9–11, 16–22). High-level abstraction is intermittently reintroduced via decompilation (Step 15) to realign instruction-level findings with function-level semantics. This interleaving of abstraction levels and repeated backward inspection exemplifies non-linear reasoning within a single tool.
We release full tool interaction traces (130–300 steps per session), with binary identifiers and confirmed vulnerability labels, to support reproducibility and independent inspection.

\section{Vulnerability Findings Statistics}
\label{app:vuln-stats}

This appendix provides detailed statistics on the vulnerability findings identified across all analysis sessions in our dataset. 
All trace analyses operate exclusively on the structure and temporal organization of reasoning-action-observation sequences.
Vulnerability characteristics are reported here solely to contextualize the corpus and to demonstrate that observed token-level implicit patterns arise under realistic security analysis conditions.

Our dataset contains 521 analysis sessions across 521 distinct binaries. Among these sessions, 198 have vulnerability analysis records, resulting in a total of 306 distinct vulnerability instances across 198 sessions that reported at least one vulnerability. 
Table~\ref{tab:vuln-overview} summarizes the distribution of vulnerability counts per session among the 198 sessions with vulnerability records. 
The 125 sessions (63.1\%) reported exactly one vulnerability. A notable subset of sessions (45, 22.7\%) reported multiple vulnerabilities, with 28 sessions reporting three or more vulnerabilities. The maximum number of vulnerabilities found in a single session is 15, all of which are CWE-121 stack-based buffer overflows.

\begin{table}[!t]
    \centering
    \small
    \caption{Vulnerability findings distribution across analysis sessions.}
    \label{tab:vuln-overview}
    \begin{tabular}{l r r}
    \toprule
    \textbf{Vulnerabilities per Session} & \textbf{Count} & \textbf{Percentage} \\
    \midrule
    1 vulnerability & 125 & 63.1\% \\
    2 vulnerabilities & 45 & 22.7\% \\
    3 vulnerabilities & 8 & 4.04\% \\
    4 vulnerabilities & 9 & 4.5\% \\
    5 vulnerabilities & 5 & 2.5\% \\
    6 vulnerabilities & 1 & 0.5\% \\
    7 vulnerabilities & 1 & 0.5\% \\
    8 vulnerabilities & 1 & 0.5\% \\
    10 vulnerabilities & 1 & 0.5\% \\
    11 vulnerabilities & 1 & 0.5\% \\
    15 vulnerabilities & 1 & 0.5\% \\
    \midrule
    \textbf{Total sessions} & \textbf{198} & \textbf{100.0\%} \\
    \bottomrule
    \end{tabular}
\end{table}

Table~\ref{tab:cwe-distribution} reports the distribution of Common Weakness Enumeration (CWE) types across all 306 vulnerability findings. The most prevalent vulnerability type is CWE-78 (OS Command Injection), accounting for 115 findings (37.6\%). This is followed by CWE-120 (Classic Buffer Overflow) with 52 findings (16.7\%) and CWE-121 (Stack-based Buffer Overflow) with 37 findings (12.2\%). Together, these top three CWE types account for 66.5\% of all identified vulnerabilities, reflecting common security issues in embedded firmware binaries.

\begin{table}[!t]
    \centering
    \small
    \caption{CWE type distribution across all vulnerability findings (Top 10).}
    \label{tab:cwe-distribution}
    \begin{tabular}{l p{4.0cm} r r}
    \toprule
    \textbf{Rank} & \textbf{CWE Type} & \textbf{Count} & \textbf{Percentage} \\
    \midrule
    1 & CWE-78 (OS Command Injection) & 115 & 37.6\% \\
    2 & CWE-120 (Classic Buffer Overflow) & 52 & 16.7\% \\
    3 & CWE-121 (Stack-based Buffer Overflow) & 37 & 12.2\% \\
    4 & CWE-134 (Format String) & 23 & 7.2\% \\
    5 & CWE-20 (Improper Input Validation) & 16 & 5.3\% \\
    6 & CWE-22 (Path Traversal) & 12 & 3.8\% \\
    7 & CWE-73 (External Control of File Name or Path) & 7 & 2.3\% \\
    8 & CWE-676 (Use of Potentially Dangerous Function) & 3 & 1.1\% \\
    9 & CWE-unknown & 2 & 0.8\% \\
    10 & CWE-123 (Write-what-where Condition) & 2 & 0.8\% \\
    \midrule
    \textbf{Top 10} & & \textbf{269} & \textbf{87.9\%} \\
    \textbf{Other } & & 37 & 12.1\% \\
    \midrule
    \textbf{Total} & & \textbf{306} & \textbf{100.0\%} \\
    \bottomrule
    \end{tabular}
\end{table}

\section{Pattern Detection Implementation}
\label{app:pattern-detection}

This section documents the concrete implementation details used to operationalize the four reasoning patterns studied in this paper. 
To avoid capturing short-lived or incidental behaviors, pattern detection requires both frequency and duration constraints. Specifically, a candidate pattern must persist across multiple non-adjacent steps and span a minimum number of reasoning turns before being recorded. Once a pattern instance is detected, the corresponding path is retired to prevent duplicate counting.
Thresholds are fixed a priori and applied uniformly across all sessions.

\subsection{Pattern 1 Detection}
\label{app:pattern-1-detection}

Pattern 1 detection operates over the \texttt{thought} field of each reasoning step, as this field captures the agent's internal reasoning process where path selection and pruning decisions are articulated. The detector implements a state-tracking algorithm that processes the trace sequentially.
 
\begin{algorithm}[!t]
\caption{Pattern 1 Detection Algorithm}
\label{alg:pattern1}
\KwIn{Sequence of reasoning steps $T = \langle s_1, s_2, \ldots, s_n \rangle$}
\KwOut{List of Pattern 1 instances $\textit{findings}$}
$\textit{pending} \leftarrow \emptyset$; $\textit{findings} \leftarrow \emptyset$\;
\For{each step $s_i \in T$}{
    $\textit{text} \leftarrow s_i.\texttt{thought}$\;
    \If{$\textit{text}$ contains revisit signal}{
        $\textit{keywords} \leftarrow \text{ExtractSemanticEntities}(\textit{text})$\;
        $\textit{pending} \leftarrow \{p \in \textit{pending} : \text{no keyword matches } p.\textit{evidence}\}$\;
    }
    \If{$\textit{text}$ matches multi-path pattern}{
        $p \leftarrow \{\textit{start\_idx}: i, \textit{evidence\_multi}: \textit{text}[:500], \textit{prune\_idx}: \text{None}\}$\;
        $\textit{pending}.\text{append}(p)$\;
    }
    \For{each $p \in \textit{pending}$ in reverse order}{
        \If{$p.\textit{prune\_idx} = \text{None}$ and $\textit{text}$ matches prune pattern}{
            $p.\textit{prune\_idx} \leftarrow i$\;
            $p.\textit{evidence\_prune} \leftarrow \textit{text}[:500]$\;
            \textbf{break}\;
        }
    }
}
\For{each $p \in \textit{pending}$}{
    \If{$p.\textit{prune\_idx} \neq \text{None}$ and $(n - p.\textit{prune\_idx}) \geq 20$}{
        $\textit{findings}.\text{append}(\{\textit{start\_step}: p.\textit{start\_idx}, \textit{prune\_step}: p.\textit{prune\_idx}, \textit{span\_after\_prune}: n - p.\textit{prune\_idx}\})$\;
    }
}
\Return{$\textit{findings}$}
\end{algorithm}

\textbf{Algorithm Overview.}
The algorithm~\ref{alg:pattern1} maintains a list of pending pattern instances throughout trace processing. For each step, it checks for revisit signals and removes pending instances whose semantic content matches the revisit keywords (lines 6--7), enforcing condition (3) of the operational definition. It then detects multi-path discussion signals and creates new pending instances (lines 8--9). Next, it checks for pruning decisions and associates them with the most recent pending instance that has not yet been pruned (lines 10--15), ensuring temporal ordering between multi-path exploration and pruning. After processing all steps, instances are recorded only if both start and prune indices are present and the span after pruning exceeds the threshold of 20 steps (lines 17--19).

The key algorithmic components are: (1) \textit{State maintenance}---the pending list tracks multiple concurrent pattern candidates; (2) \textit{Revisit exclusion}---semantic entity extraction enables precise filtering when agents explicitly return to previously abandoned paths; (3) \textit{Temporal ordering enforcement}---pruning decisions are associated with the most recent unpruned instance; and (4) \textit{Threshold filtering}---the 20-step minimum span ensures detected instances reflect sustained behavioral focus rather than transient narrowing behaviors.

\subsection{Pattern 2 Detection}
\label{app:pattern-2-detection}

Pattern 2 detection operates over the \texttt{thought} field of each reasoning step, tracking path selection decisions and their subsequent influence on analysis behavior. The detector implements a state-tracking algorithm that maintains active path candidates and monitors continuation, alternative mentions, and contradiction absorption patterns.

\begin{algorithm}[!t]
\caption{Pattern 2 Detection Algorithm}
\label{alg:pattern2}
\KwIn{Sequence of reasoning steps $T = \langle s_1, s_2, \ldots, s_n \rangle$}
\KwOut{List of Pattern 2 instances $\textit{findings}$}
$\textit{paths} \leftarrow \emptyset$; $\textit{findings} \leftarrow \emptyset$\;
\For{each step $s_i \in T$}{
    $\textit{text} \leftarrow s_i.\texttt{thought}$; $\textit{kw} \leftarrow \text{ExtractSemanticEntities}(\textit{text})$\;
    \If{$\textit{text}$ contains global re-evaluation signal}{
        $\textit{paths} \leftarrow \{p \in \textit{paths} : \text{no keyword matches } p.\textit{kw}\}$\;
    }
    \If{$\textit{text}$ matches path-selection pattern}{
        $p \leftarrow \{\textit{start}: i, \textit{kw}: \textit{kw}, \textit{snippet}: \textit{text}[:500], \textit{cnt\_cont}: 0, \textit{cnt\_alt}: 0, \textit{cnt\_contrad}: 0\}$; $\textit{paths}.\text{append}(p)$\;
    }
    \For{each $p \in \textit{paths}$}{
        \If{$\textit{kw}$ overlaps with $p.\textit{kw}$}{
            $p.\textit{cnt\_cont} \leftarrow p.\textit{cnt\_cont} + 1$; $p.\textit{last\_cont} \leftarrow i$\;
        }
        \If{$\textit{text}$ matches alternative-mentioned pattern and keywords overlap}{
            $p.\textit{cnt\_alt} \leftarrow p.\textit{cnt\_alt} + 1$\;
        }
        \If{$\textit{text}$ matches contradiction-absorbed pattern and keywords overlap}{
            $p.\textit{cnt\_contrad} \leftarrow p.\textit{cnt\_contrad} + 1$; $p.\textit{last\_contrad} \leftarrow i$\;
        }
    }
    \For{each $p \in \textit{paths}$}{
        \If{$p.\textit{cnt\_cont} \geq 5$ and span $\geq 10$}{
            $\textit{end} \leftarrow \max(p.\textit{last\_cont}, p.\textit{last\_contrad}, p.\textit{start})$\;
            $\textit{findings}.\text{append}(\{\textit{start\_step}: p.\textit{start}, \textit{end\_step}: \textit{end}, \textit{span}: \textit{end} - p.\textit{start}, \textit{cont\_count}: p.\textit{cnt\_cont}, \textit{alt\_count}: p.\textit{cnt\_alt}, \textit{contrad\_count}: p.\textit{cnt\_contrad}\})$; $\textit{paths}.\text{remove}(p)$\;
        }
    }
}
\Return{$\textit{findings}$}
\end{algorithm}

\textbf{Algorithm Overview.}
The algorithm~\ref{alg:pattern2} maintains a list of active path selections throughout trace processing. For each step, it checks for global re-evaluation signals and removes active paths whose semantic content matches the re-evaluation keywords (lines 6--7), excluding episodes that exhibit deliberate strategy revision. It then detects path-selection decisions and creates new active path instances (lines 7--8). For each active path, the algorithm tracks continuation through semantic entity overlap (lines 9--10), monitors alternative mentions within the same context (line 11), and detects contradiction absorption patterns (lines 12--13). Instances are recorded when continuation count reaches at least 5 occurrences and the span exceeds 10 steps (lines 19--23).

The key algorithmic components are: (1) \textit{Semantic entity extraction}---function names, memory addresses, and path-related terms are extracted to track continuity; (2) \textit{Path continuation tracking}---overlapping semantic entities between steps indicate sustained focus; (3) \textit{Alternative mention detection}---signals of alternative paths mentioned but not explored are tracked; (4) \textit{Contradiction absorption detection}---linguistic patterns indicating acknowledgment of contradictions without triggering global re-evaluation; and (5) \textit{Global re-evaluation exclusion}---explicit signals of approach switching remove affected paths.

\subsection{Pattern 3 Detection}
\label{app:pattern-3-detection}

Pattern 3 detection operates over the \texttt{thought} field of each reasoning step, tracking deferred path mentions, development events, and targeted backtracking. The detector implements a state-tracking algorithm that maintains lists of deferred and developed paths, monitoring for backtracking signals that indicate revisitation of previously deferred candidates.

\begin{algorithm}[!t]
\caption{Pattern 3 Detection Algorithm}
\label{alg:pattern3}
\KwIn{Sequence of reasoning steps $T = \langle s_1, s_2, \ldots, s_n \rangle$}
\KwOut{List of Pattern 3 instances $\textit{findings}$}
$\textit{deferred} \leftarrow \emptyset$; $\textit{materialized} \leftarrow \emptyset$; $\textit{findings} \leftarrow \emptyset$\;
\For{each step $s_i \in T$}{
    $\textit{text} \leftarrow s_i.\texttt{thought}$; $\textit{kw} \leftarrow \text{ExtractSemanticEntities}(\textit{text})$\;
    \If{$\textit{text}$ matches multi-candidates pattern}{
        $p \leftarrow \{\textit{m\_idx}: i, \textit{kw}: \textit{kw}, \textit{snippet}: \textit{text}[:500], \textit{mat}: \text{False}, \textit{bt}: \text{False}\}$; $\textit{deferred}.\text{append}(p)$\;
    }
    \If{$\textit{text}$ matches development pattern}{
        \For{each $p \in \textit{deferred}$}{
            \If{$\textit{kw}$ overlaps with $p.\textit{kw}$}{
                $p.\textit{mat} \leftarrow \text{True}$; $p.\textit{mat\_idx} \leftarrow i$; $\textit{materialized}.\text{append}(\{\textit{deferred\_idx}: p.\textit{m\_idx}, \textit{mat\_idx}: i, \textit{kw}: \textit{kw}\})$; \textbf{break}\;
            }
        }
    }
    \If{$\textit{text}$ matches backtracking pattern}{
        $\textit{new\_sig} \leftarrow \text{CheckNewSignal}(\textit{text}, T, i)$; $\textit{impasse} \leftarrow \text{CheckImpasse}(\textit{text}, T, i)$; $p \leftarrow \text{FindMatchingDeferredPath}(\textit{deferred}, \textit{kw}, i)$\;
        \If{$p \neq \text{None}$}{
            $p.\textit{bt} \leftarrow \text{True}$; $p.\textit{bt\_idx} \leftarrow i$; $p.\textit{bt\_snippet} \leftarrow \textit{text}[:500]$\;
        }
    }
}
\For{each $p \in \textit{deferred}$}{
    \If{$p.\textit{bt}$ and $(p.\textit{bt\_idx} - p.\textit{m\_idx}) \geq 1$}{
        $\textit{mat\_idx} \leftarrow \text{FindMatBeforeBt}(\textit{materialized}, p.\textit{bt\_idx})$\;
        $\textit{findings}.\text{append}(\{\textit{deferred\_step}: p.\textit{m\_idx}, \textit{mat\_step}: \textit{mat\_idx}, \textit{bt\_step}: p.\textit{bt\_idx}, \textit{span}: p.\textit{bt\_idx} - p.\textit{m\_idx}\})$\;
    }
}
\Return{$\textit{findings}$}
\end{algorithm}

\textbf{Algorithm Overview.}
The algorithm~\ref{alg:pattern3} maintains two lists: deferred paths (lightweight mentions) and developed paths (explicitly developed). For each step, it detects multi-candidate mentions and creates deferred path records (lines 7--8), identifies development events by matching semantic entities between current text and deferred paths (lines 9--14), and detects targeted backtracking through linguistic patterns indicating explicit return to previously mentioned paths (lines 15--19). Backtracking detection includes checks for new evidence signals or analysis impasses. Instances are recorded when a deferred path is backtracked after at least one intervening step (lines 21--24).

The key algorithmic components are: (1) \textit{Deferred path tracking}---multiple candidate paths are recorded when mentioned without full expansion; (2) \textit{Development detection}---semantic entity overlap indicates explicit development; (3) \textit{Targeted backtracking identification}---linguistic patterns combined with semantic entity matching distinguish explicit revisitation from simple continuation; (4) \textit{Context signal detection}---new evidence or impasse signals provide additional context for revisitation triggers; and (5) \textit{Temporal separation enforcement}---minimum span requirements ensure backtracking represents revisitation rather than immediate continuation.

\subsection{Pattern 4 Detection}
\label{app:pattern-4-detection}

Pattern 4 detection operates over the \texttt{thought} field of each reasoning step, identifying path-selection or prioritization decisions and their justifications. The detector implements a feature-based classification algorithm that evaluates multiple linguistic signals to determine whether a step represents knowledge-guided prioritization.

\begin{algorithm}[!t]
\caption{Pattern 4 Detection Algorithm}
\label{alg:pattern4}
\KwIn{Sequence of reasoning steps $T = \langle s_1, s_2, \ldots, s_n \rangle$}
\KwOut{List of Pattern 4 instances $\textit{findings}$}
$\textit{findings} \leftarrow \emptyset$\;
\For{each step $s_i \in T$}{
    $\textit{text} \leftarrow s_i.\texttt{thought}$; $\textit{kw} \leftarrow \text{ExtractSemanticEntities}(\textit{text})$\;
    $\textit{analogy} \leftarrow \text{MatchPattern}(\textit{text}, \text{analogy\_pat})$\;
    $\textit{priority} \leftarrow \text{MatchPattern}(\textit{text}, \text{priority\_pat})$\;
    $\textit{sig\_score} \leftarrow \text{MatchPattern}(\textit{text}, \text{sig\_score\_pat})$\;
    $\textit{partial} \leftarrow \text{MatchPattern}(\textit{text}, \textit{partial\_pat})$\;
    $\textit{justif} \leftarrow \text{MatchPattern}(\textit{text}, \text{justif\_pat})$\;
    $\textit{select} \leftarrow \text{MatchPattern}(\textit{text}, \text{select\_pat})$\;
    $\textit{feat\_cnt} \leftarrow \text{CntFeat}(\textit{analogy}, \textit{priority}, \textit{sig\_score},$\\
    \phantom{$\textit{feat\_cnt} \leftarrow \text{CntFeat}($}$\textit{partial}, \textit{justif}, \textit{select})$\;
    $\textit{core} \leftarrow \textit{analogy} \text{ or } \textit{sig\_score}$\; 
    $\textit{decision} \leftarrow \textit{select} \text{ or } \textit{priority}$\;
    \If{$(\textit{core} \text{ and } \textit{decision}) \text{ or } (\textit{feat\_cnt} \geq 3 \text{ and } \textit{core})$}{
        $\textit{justif\_snip} \leftarrow \text{ExtractJustificationSnippet}(\textit{text})$\;
        $\textit{findings}.\text{append}(\{\textit{step}: i, \textit{feat\_cnt}: \textit{feat\_cnt}, \textit{justif\_snip}: \textit{justif\_snip}, \textit{kw}: \textit{kw}\})$\;
    }
}
\Return{$\textit{findings}$}
\end{algorithm}

\textbf{Algorithm Overview.}
The algorithm~\ref{alg:pattern4} evaluates each reasoning step for six linguistic features: analogy signals, priority assignments, signal scoring, partial evidence references, justification statements, and path-selection actions. A Pattern 4 instance is recorded when either (1) a core feature (analogy or signal scoring) co-occurs with a decision feature (path selection or priority assignment), or (2) at least three features are present including a core feature (lines 14--16). This two-tier criterion ensures that detected instances represent reasoned prioritization decisions rather than incidental mentions.

The key algorithmic components are: (1) \textit{Multi-feature detection}---six complementary linguistic patterns capture different aspects of prioritization reasoning; (2) \textit{Core feature requirement}---analogy or signal scoring must be present; (3) \textit{Decision feature requirement}---path selection or priority assignment must be present; (4) \textit{Feature count threshold}---alternative criterion of three or more features with a core feature provides robustness; and (5) \textit{Justification extraction}---snippets containing decision rationale are extracted for evidence preservation.

\section{Metric Measurement Details}
\label{app:metric}

All metrics are computed over pattern-aligned reasoning segments extracted from full interaction traces.
For each detected pattern instance, we identify the corresponding contiguous segment of assistant steps and compute the following metrics using deterministic, script-based procedures released in the artifact.

\textbf{Path Length ($L$)}.
The total number of assistant steps within the pattern-aligned segment, counted from the segment start to end.
This metric directly measures the reasoning horizon associated with each pattern instance.

\textbf{Forward Step Ratio ($F$)}.
The fraction of assistant steps labeled with \texttt{status="continue"} among all assistant steps in the segment.
This captures the degree of forward progression relative to halting or revision behavior.

\textbf{Branching Factor ($B$)}.
Branching behavior is operationalized by detecting explicit decision points in the \texttt{thought} field using keyword-based matching (e.g., ``choose'', ``select'', ``decide'').
At each decision point, we detect references to multiple alternatives using predefined linguistic patterns (e.g., ``multiple paths'', ``several options'', ``various candidates'').
The branching factor is estimated as
$B = 1.0 + \alpha \cdot r$,
where $r$ is the fraction of decision points mentioning multiple alternatives and $\alpha = 1.5$ is a fixed scaling constant.
This heuristic provides a consistent approximation of how many alternatives are considered during reasoning, without assuming explicit control-flow reconstruction.

\textbf{Backtrack Count ($R$)}.
Backtracking events are detected by matching explicit revision language in the \texttt{thought} field (e.g., ``backtrack'', ``go back'', ``return to'', ``revisit'') as well as references to earlier analysis locations.
For Pattern~3 instances, $R$ is set to 1 by definition, as Pattern~3 represents targeted backtracking behavior by construction.

\textbf{Pruning Rate ($P$)}.
Pruning is measured at detected decision points.
We identify pruning behavior by matching elimination-related expressions (e.g., ``skip'', ``ignore'', ``discard'', ``not worth'', ``irrelevant'', ``focus on'').
The pruning rate is defined as
$$P = \frac{\text{number of decision points with pruning signals}}{\text{total number of decision points}}$$
This metric quantifies how aggressively candidate paths are eliminated during reasoning.

\subsection{Tool Usage Topology Metrics}

For each pattern-aligned segment, we extract the sequence of tool invocations from assistant steps and compute the following topology metrics.

\textbf{Command Diversity}.
The number of distinct command types appearing in the segment.
Command types are extracted by parsing the \texttt{action} field and retaining only the base command name (parameters removed).

\textbf{Sequence Length}.
The total number of tool invocations in the segment, measured as the count of assistant steps with non-empty \texttt{action} fields.

\textbf{Max Depth}.
The maximum length of consecutive repetitions of the same command type within the invocation sequence.
This captures repetitive tool usage patterns.

\textbf{Max Fan-out}.
The maximum number of distinct successor command types following any given command.
We construct a directed transition graph over command types and compute fan-out per node.

\textbf{Cycle Presence}.
A binary indicator denoting whether the command transition graph contains at least one directed cycle, detected via depth-first search.

\textbf{Transition Entropy}.
For each command $c_i$, we compute the entropy of its outgoing transition distribution
$$H(c_i) = -\sum_{c_j} p(c_j|c_i)\log_2 p(c_j|c_i)$$
Overall transition entropy is computed as a weighted average over all commands, weighted by their transition frequencies.
Lower values indicate deterministic tool usage, while higher values indicate diversified transitions.

All metrics are computed using fixed scripts provided in the artifact repository, enabling full reproducibility across binaries, patterns, and experimental settings.


\end{document}